\pdfoutput=1

\documentclass[11pt]{article}

\usepackage[]{latex/acl}

\usepackage{times}
\usepackage{latexsym}

\usepackage[T1]{fontenc}

\usepackage[utf8]{inputenc}

\usepackage{microtype}

\usepackage{inconsolata}

\usepackage{multirow}
\usepackage{booktabs}
\usepackage{graphicx}
\usepackage{svg}
\usepackage{xcolor}
\usepackage{color}
\usepackage{colortbl}
\usepackage{amsfonts}
\usepackage{subcaption}
\usepackage{amsmath}
\usepackage{dsfont}
\usepackage{bbold}

\title{Incremental Sequence Labeling: A Tale of Two Shifts}

\author{Shengjie Qiu, Junhao Zheng, Zhen Liu, Yicheng Luo, Qianli Ma* \\
  School of Computer Science and Engineering, \\
  South China University of Technology, Guangzhou, China\\
  \texttt{shengjieqiu6@gmail.com},
  \texttt{junhaozheng47@outlook.com}, \\
  \texttt{\{cszhenliu,csluoyicheng\}@mail.scut.edu.cn},
  \texttt{qianlima@scut.edu.cn}\thanks{*Corresponding author}}

\begin{document}
\maketitle
\begin{abstract}
The incremental sequence labeling task involves continuously learning new classes over time while retaining knowledge of the previous ones. Our investigation identifies two significant semantic shifts: E2O (where the model mislabels an old entity as a non-entity) and O2E (where the model labels a non-entity or old entity as a new entity). Previous research has predominantly focused on addressing the E2O problem, neglecting the O2E issue. This negligence results in a model bias towards classifying new data samples as belonging to the new class during the learning process.
To address these challenges, we propose a novel framework, \textbf{I}ncremental \textbf{S}equential \textbf{L}abeling without \textbf{S}emantic \textbf{S}hifts (IS3). Motivated by the identified semantic shifts (E2O and O2E), IS3 aims to mitigate catastrophic forgetting in models. As for the E2O problem, we use knowledge distillation to maintain the model's discriminative ability for old entities. Simultaneously, to tackle the O2E problem, we alleviate the model's bias toward new entities through debiased loss and optimization levels.
Our experimental evaluation, conducted on three datasets with various incremental settings, demonstrates the superior performance of IS3 compared to the previous state-of-the-art method by a significant margin. 
The data, code, and scripts are publicly available \footnote{\href{https://github.com/zzz47zzz/codebase-for-incremental-learning-with-llm}{https://github.com/zzz47zzz/codebase-for-incremental-learning-with-llm}}.
\end{abstract}

\section{Introduction}
\label{sec:introduction}
The conventional sequence labeling task typically involves categorizing data into a predetermined set of fixed categories \cite{lample2016neural}. However, this approach may need to be revised in natural language processing scenarios, such as the named entity recognition task, where new types of entities continuously emerge. Adapting a fixed set of categories becomes challenging when faced with the dynamic nature of new entity classification requirements. Consequently, continuous model updates are essential to accommodating evolving entity types. Previous studies have advocated for adopting continual learning \cite{parisi2019continual, monaikul2021continual}, also known as lifelong learning or incremental learning. Continual learning is a paradigm designed to train models capable of adapting to the continual addition of new categories in real-world scenarios while ensuring that knowledge of old categories is retained. For instance, voice assistants like Siri frequently encounter new event types, such as pandemics, need to understand and provide health-protecting information based on the user's latest intent. \cite{monaikul2021continual}.

\begin{figure}[t]
\centering
\includegraphics[width=1\linewidth]{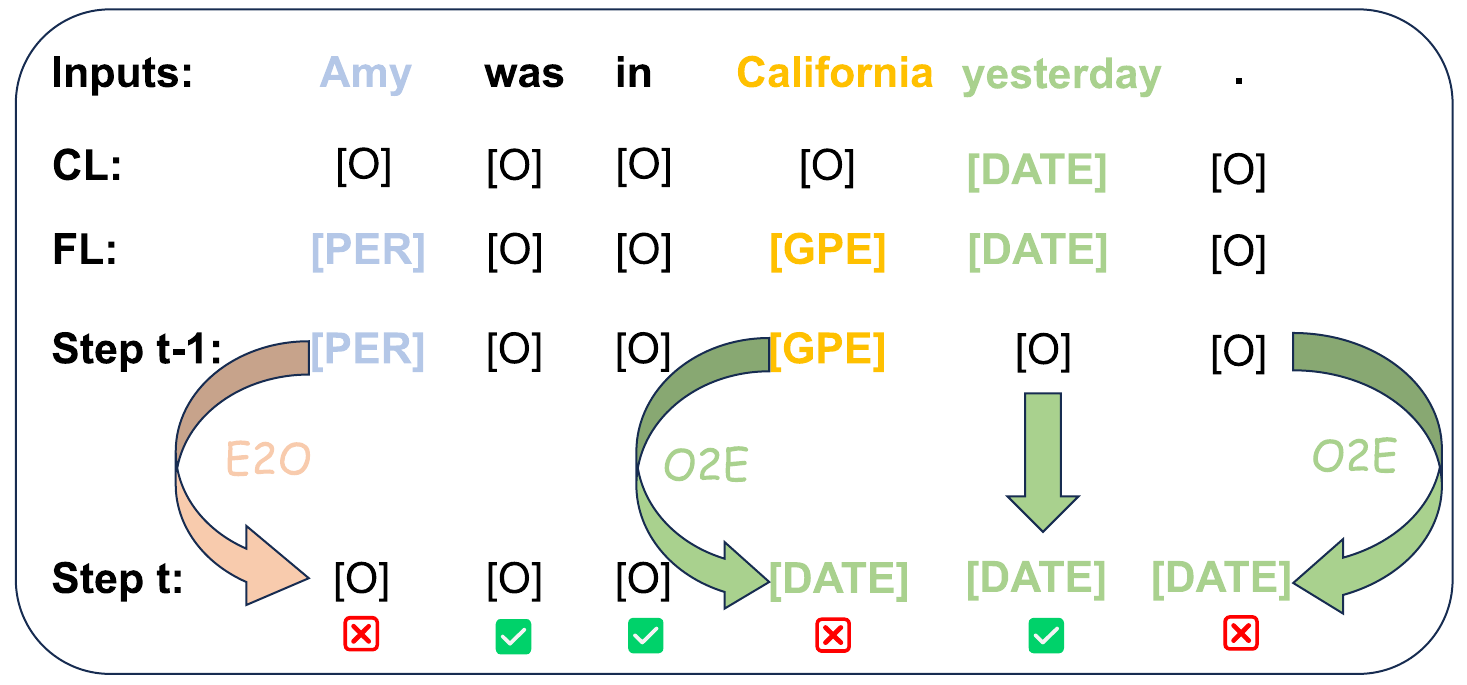}
\caption{A sample shows two shifts in incremental sequence labeling. E2O denotes the semantic shift of an old entity (such as [PER]) to a non-entity ([O]), and O2E denotes the semantic shift of a non-entity ([O]) or an old entity(such as [GPE]) to a new entity (such as [DATE]). \textbf{Inputs} means input sentence. \textbf{CL} means \textit{current ground-truth label} at step $t$. \textbf{FL} means the \textit{full ground-truth label} for all steps. \textbf{Step $t-1$ and Step $t$} means the predictions in step $t-1$ and $t$.}
\label{fig:mot}
\end{figure}
Due to constraints imposed by storage limitations and privacy concerns, there exists a shortage of training data about the old categories \cite{he2022exemplar}. Additionally, the manual relabeling of all categories within the new training dataset would incur substantial costs and time investment \cite{de2021continual, bang2021rainbow}. Consequently, the model undergoes continuous updates using a freshly acquired dataset comprising the new categories. As depicted in Figure \ref{fig:mot}, the model undergoes training based on the \textit{current ground-truth label} and undergoes testing using the \textit{full ground-truth label}. 

The incremental sequence labeling task faces a significant challenge known as the catastrophic forgetting problem, as extensively discussed in previous studies \cite{mccloskey1989catastrophic, robins1995catastrophic, goodfellow2013empirical, kirkpatrick2017overcoming,zheng2024concept}. This issue manifests as semantic shifts, leading to a decrease in the discriminative power of entity classes \cite{zhang2023task, ma-etal-2023-learning}. In this paper, we decompose the problem into two primary semantic shifts in the incremental sequence task: E2O and O2E.
The first semantic shift, E2O, arises from the presence of non-entities, potential old entities (mislabelled as non-entities), and new entities in the new dataset. Progress has been made in addressing E2O through methods falling into three categories:
(1) Methods based on knowledge distillation: For instance, RDP proposes a knowledge distillation loss incorporating inter-task relations \cite{zhang2023task}. At the same time, CFPD introduces a pooled feature distillation loss to alleviate catastrophic forgetting \cite{zhang2023continual}.
(2) Methods based on pseudo-labels: OCILNER utilizes class prototypes to label new data \cite{ma-etal-2023-learning}, and CPFD employs old models to label predictions of new data.
(3) Methods based on freezing models: Examples include ICE \cite{liu-huang-2023-teamwork}, which freezes the backbone model and old classifiers to maintain the stability of the old classes at the expense of learning new classes.

Existing methods primarily focus on addressing the E2O shift, neglecting the bias towards the emergence of new classes and the consequential second semantic shift, O2E. To address both semantic shifts, we propose a novel framework called \textbf{I}ncremental \textbf{S}equential \textbf{L}abeling without \textbf{S}emantic \textbf{S}hifts (IS3).
IS3 consists of two key components: First, we apply the knowledge distillation method to tackle the E2O shift. Second, we address the O2E shift on two fronts. At the loss function level, we introduce a debiased cross-entropy loss function to mitigate the model's impact on old class distributions, reducing its inclination towards new entities. At the optimization level, we introduce a prototype-based approach to balance the imbalanced contributions of old and new entities during batch updates, which aims to increase the involvement of old entities in the optimization process. Importantly, IS3 adopts a storage-efficient approach, maintaining only one prototype per class with minimal storage costs. Class feature centers serve as prototypes, ensuring no direct correspondence to actual sample information and mitigating privacy leakage concerns.

The contribution of our work can be summarized as follows:
\begin{itemize}
    \item We propose a novel perspective on the semantic shift problem in incremental sequence labeling task by categorizing the catastrophic forgetting problem into E2O and O2E.
    \item We propose a novel framework, \textbf{I}ncremental \textbf{S}equential Labeling without \textbf{S}emantic \textbf{S}hifts (IS3), to solve the two semantic shifts simultaneously.
    \item We conduct experiments under nine CIL settings on three datasets, and our method outperforms the previous state-of-the-art methods.
\end{itemize}

\section{Related Work}
\label{sec:related Work}
\textbf{Incremental Learning}\quad The model continually acquires new tasks intending to achieve optimal performance on tasks previously learned \cite{gepperth2016incremental,wu2019large,van2022three,zheng2023learn}. There are three main categories of current incremental learning methods: regularization-based, rehearsal-based, and architecture-based. Regularization-based methods place constraints on model weights \cite{kirkpatrick2017overcoming,zenke2017continual}, representations of intermediate layer features \cite{hou2019learning,douillard2020podnet}, and output probabilities \cite{li2017learning,zheng-etal-2023-preserving}. Rehearsal-based methods overcome forgetting by saving some of the data containing the old classes for learning with the new classes \cite{lopez2017gradient,shin2017continual}. Alternatively, architecture-based approaches involve dynamically expanding the network structure to allow for more data as new classes are added \cite{hou2018lifelong,yan2021dynamically}.\\
\textbf{Incremental Sequence Labeling}\quad The traditional sequence labeling task is the task of labeling each token of a one-dimensional linear input sequence, which requires each token to be categorized according to its contextual content\cite{rei2016attending,akbik2018contextual}. However, previous methods can only recognize classes in a fixed set. Therefore, continuous learning paradigms are introduced in sequence labeling tasks, including incremental named entities \cite{monaikul2021continual,zheng-etal-2022-distilling,zhang2023continual}, incremental event detection \cite{cao2020incremental,yu2021lifelong}, and so on.

\begin{figure}[t]
\centering
    \includegraphics[width=0.96\linewidth]{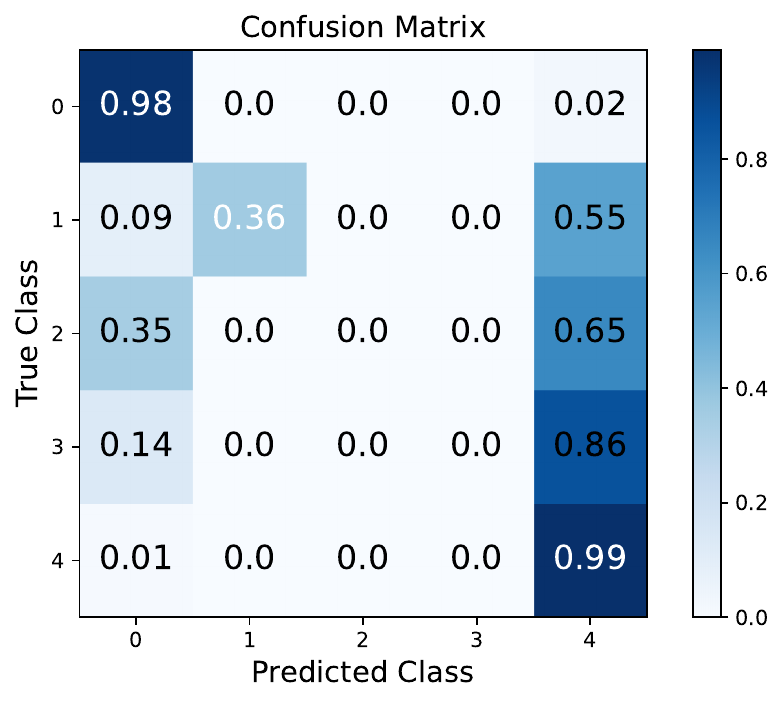}
    \caption{Confusion Matrix of the ExtendNER method in Task 4. It indicates that the model predicts the old entities as new entities with high probability and predicts the old entity as non-entity, with severe O2E semantic shift and E2O semantic shift.}
    \label{fig:confusion matrix}
\end{figure}
Methods for incremental sequence labeling tasks can be categorized into distillation-based, rehearsal-based, and other approaches. Distillation-based methods encompass ExtendNER \cite{monaikul2021continual}, which is the pioneer in applying knowledge distillation to incremental sequence labeling task, RDP \cite{zhang2023task} with a relational distillation approach, and CPFD \cite{zhang2023continual} utilizing pooled features distillation loss. CFNER \cite{zheng-etal-2022-distilling} introduces a causal framework for extracting new causal effects in entities and non-entities.
Rehearsal-based approaches include KCN \cite{cao2020incremental} and KD+R+K \cite{yu2021lifelong}, both employing rehearsal samples to address the class imbalance and catastrophic forgetting in incremental event detection. L\&R \cite{xia2022learn} proposes a learn-and-review framework by training a new backbone model and a generative model simultaneously, generating synthetic samples of the old class to be trained with new samples. OCILNER \cite{ma-etal-2023-learning} uses rehearsal samples to compute class feature centers as class prototypes, generates an entity-oriented feature space through comparative learning, and annotates new data with pseudo-labels using class prototypes.
Other methods encompass span-based and freezing model-based approaches, among others.

The mentioned methodologies primarily focus on preserving the existing knowledge of the model and do not explicitly consider the implications of transitioning from non-entity to entity semantics. In contrast, our proposed method, IS3, provides a fresh perspective on model forgetting by addressing the model's inclination towards new classes during task adaptation. IS3 not only addresses issues related to model mislabeling, indirectly mitigating the problem of semantic migration from entity to non-entity, but also handles the challenge of semantic migration from non-entity to entity. By recognizing and addressing the model's bias towards new classes during adaptation, our approach offers a comprehensive solution to the dynamic challenges associated with transitioning between different semantic categories.
\section{Problem Formulation}
\label{sec:problem formulation}
Formally, the objective of incremental sequence labeling is to acquire knowledge through a series of tasks $\mathcal{T} = \{\mathcal{T}_1, \mathcal{T}_2,\ldots,\mathcal{T}_N\}$. Each task contains its dataset $\mathcal{D}_t = \{(x^i,y^i)| y^i\in \mathcal{Y}_t\}$ where $(x^i,y^i)$ is a pair formed by the input token sentence and the label corresponding to each token in the sentence and $\mathcal{Y}_t$ stands for the current label set. Notably, $y^i$ only labels the token corresponding to the current task $t$, and the other tokens are labeled as O class (potential old entities $\mathcal{Y}_{1:t}$, and unseen entities $\mathcal{Y}_{t+1:N}$). At task $t$ ($t>1$), the new model $\mathcal{M}_t$ learns only from the new dataset and is expected to perform well on the learned classes $\bigcup_{i=1}^t\mathcal{Y}_i$.
\section{Method}
\label{sec:method}
\begin{figure*}[t]
\centering
    \includegraphics[width=0.98\linewidth]{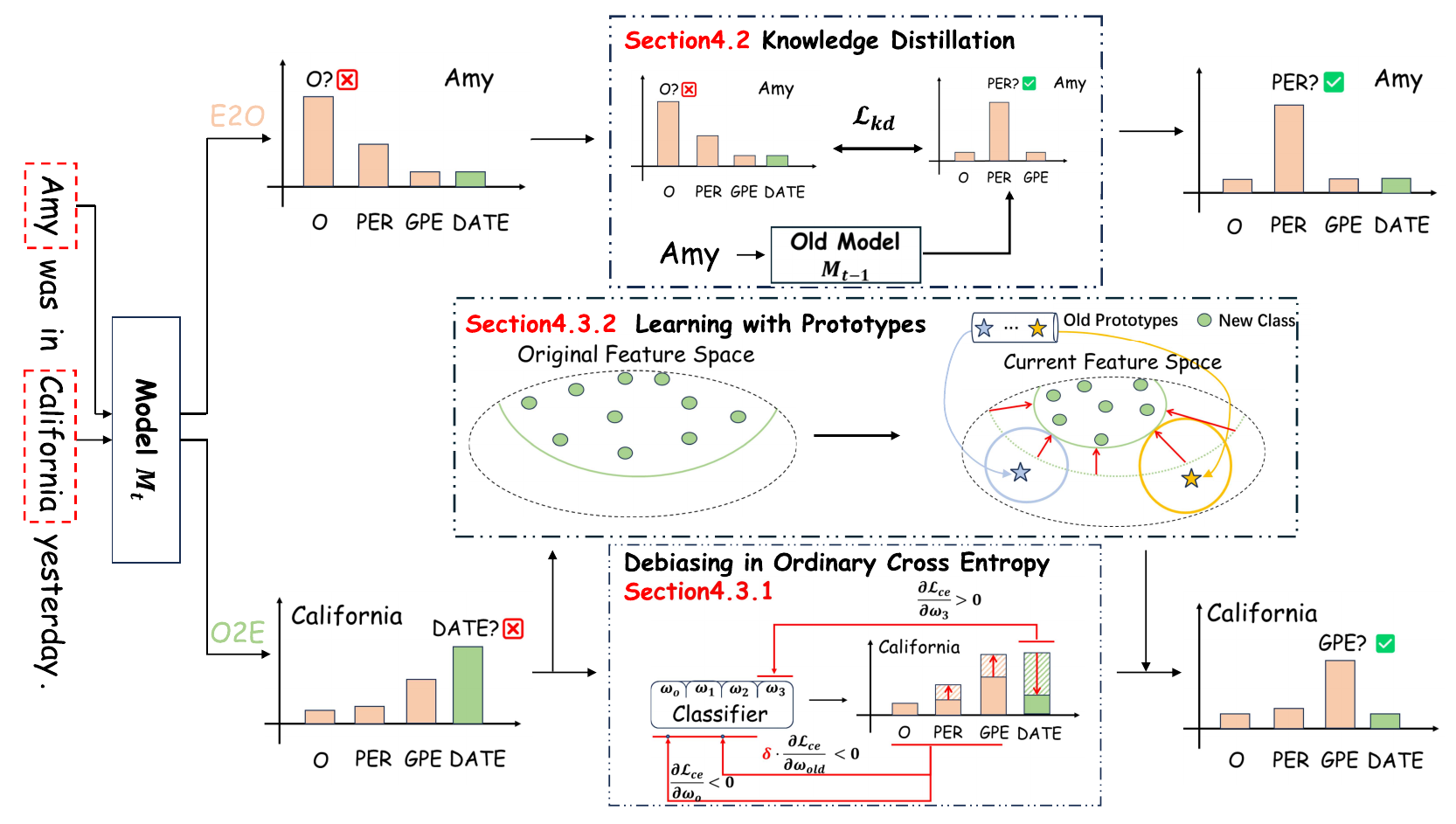}
    \caption{Overview of our framework IS3 for incremental sequence labeling. We solve the O2E problem by distillation loss $L_{kd}$. Besides, we use two modules: debiased cross-entropy loss $L_{ce}^{Debias}$ and prototype learning to solve the E2O problem.}
    \label{fig:model}
\end{figure*}

In this section, we systematically address the catastrophic forgetting problem by decomposing it into two distinct semantic shift challenges (Section \ref{sec:two shifts}). Subsequently, we present a comprehensive framework designed to address these semantic shifts individually, focusing on E2O in Section \ref{sec:E2O} and O2E in Section \ref{sec:O2E}. The overarching goal is to effectively mitigate the catastrophic forgetting problem, as illustrated in Figure \ref{fig:model}.

\subsection{Two semantic shift problems}
\label{sec:two shifts}
In the incremental sequence labeling task, semantic shift can be decomposed into entity to non-entity semantic shift and non-entity to entity semantic shift, which are abbreviated as E2O and O2E.

\begin{figure}[t]
\centering
    \includegraphics[width=0.98\linewidth]{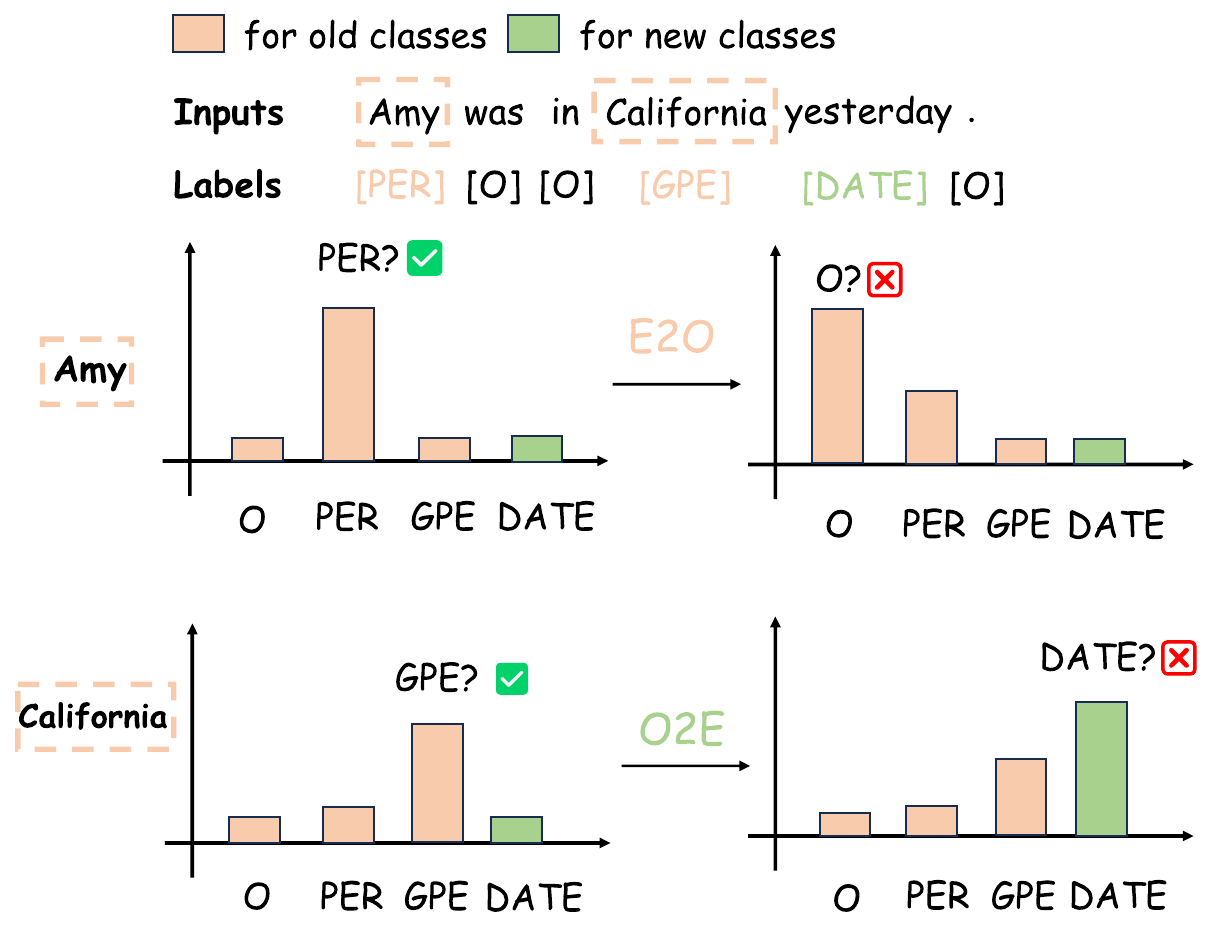}
    \caption{Illustration of E2O and O2E. When "Amy" encounters E2O problem, the label is biased from [PER] to [O]. "California" encounters O2E problem, the label is shifted from [GPE] to [DATE].}
    \label{fig:mot2}
\end{figure}

\textbf{E2O} refers to the model incorrectly categorizing entities as non-entities during the learning process. This misclassification stems from the incremental sequence labeling task, where only new entities are labeled in the new dataset, potentially causing old entities to be erroneously labeled as non-entities. For instance, in Figure \ref{fig:mot2}, the name "Amy" is mistakenly labeled as a non-entity. This misclassification induces a gradual shift in the semantics of old entities towards non-entities, leading to a blurred boundary between the two classes.
Several previous approaches have addressed this bias issue. Methods like RDP focus on designing improved distillation techniques to maintain the stability of the model's old entities. Similarly, OCILNER utilizes comparative learning to obtain a more discriminative feature space, clarifying the classification boundaries between entities and non-entities. These strategies aim to mitigate the impact of E2O, ensuring a more accurate preservation of entity semantics during incremental sequence labeling tasks.

\textbf{O2E} signifies the model incorrectly labeling non-entities or old entities as new entities during the learning process. As seen in Figure \ref{fig:mot2}, our observations indicate that the model maintains good discrimination between old entities. However, Figure \ref{fig:confusion matrix} shows a bias towards new entities in predictions during incremental learning. Our research identifies two key contributing factors to this bias.

The first factor is related to the classifier dimension's predisposition. When learning new entities, the ordinary cross-entropy function induces the model to fit and converge faster on the distribution of new entities by excessively penalizing the classifier dimension associated with old entities. This over-penalization of old entities results in a pronounced bias in significant classification scores towards the new classes.

The second factor involves a tendency at the feature optimization level. The current dataset mainly contains samples of new entities with minimal representation from other entities, including potential old and future new entities, to facilitate effective learning of new entities. As a result, in the same batch, the probability of old entities participating in model optimization is much lower than the probability of new entities' participation. Consequently, there is a predisposition towards new categories at the feature optimization level. Addressing these aspects is crucial for mitigating the O2E semantic shift and achieving more balanced and accurate predictions during incremental sequence labeling tasks.

Notably, the E2O and O2E problems are interconnected. If the O2E problem occurs in the model during incremental sequence labeling, it can gradually blur the boundaries between entities and entities and among entities. It can also indirectly contribute to the E2O problem, ultimately impacting the model's discriminative ability. We will address these two semantic biases separately to mitigate catastrophic forgetting during incremental sequence labeling.
\subsection{Solving E2O problem via knowledge distillation}
\label{sec:E2O}
When learning the current task $t$, the model $\mathcal{M}^t$ is trained on the training examples with the current entities, which often leads to catastrophic forgetting for the old entities. To alleviate the E2O problem, we use knowledge distillation \cite{hinton2015distilling}. This method preserves the prior knowledge by distilling the output probabilities from the old $\mathcal{M}^{t-1}$ to the current model $\mathcal{M}^t$. Therefore, the objective function for solving the E2O problem can be expressed as:
\begin{equation}
\mathcal{L}_{kd} = \frac{1}{|\mathcal{D}_t|}\sum_{i=1}^{|\mathcal{D}_t|} \hat{y}^{t-1}_i \log\hat{y}^t_i,
\label{equ:Loss_kl}
\end{equation}
where $\hat{y}^{t-1}_i$ and $\hat{y}^t_i$ represent the output probabilities of the current model and the old model respectively. 

Through Eq.\ref{equ:Loss_kl}, "Amy" in Figure \ref{fig:model} corrects the current model's incorrect labeling via the output probabilities provided by the old model, thus maintaining discriminative properties between old entities.
\subsection{Solving O2E problem}
\label{sec:O2E}
In this section, we address the O2E problem at the debiased loss and feature optimization levels.
\subsubsection{Debiasing in Ordinary Cross Entropy}
The overall model parameters are defined as $\Theta = \{\theta, \omega\}$. The model's backbone $f_\theta:X\rightarrow\mathbb{R}^d$ extracts feature embeddings of dimension $d$ from the inputs. Following the backbone, a linear classifier produces logits $\Phi(\cdot) = \omega^T\cdot f_\theta(\cdot):X\rightarrow\mathbb{R}^{\left| \mathcal{Y}_t \right|}$, where $\omega$ represents the classifier weights for the corresponding dimensions. As the number of classes of recognizable entities increases as well, the dimension of the classifier increases. The model is trained by a cross-entropy loss function, which is defined as:
\begin{equation}
\begin{aligned}
\mathcal{L}_{ce} &= -\frac{1}{|\mathcal{D}_t|}\sum_{i=1}^{|\mathcal{D}_t|} y_i \log\left(\frac{e^{\Phi_{y_i} (x_i)}}{\sum_{y^\prime \in \mathcal{Y}_t}  e^{\Phi_{y^\prime}}(x_i)} \right) \\
&= \frac{1}{|\mathcal{D}_t|}\sum_{i=1}^{|\mathcal{D}_t|} \log [1+\sum_{y\prime\neq y_i}e^{\Phi_{y\prime}(x_i)-\Phi_{y_i}(x_i)}], \\
\end{aligned}
\label{equ:Loss_ce}
\end{equation}
where $y_i$ denotes the label of the new entity for the current incremental step $t$. Figure \ref{fig:confusion matrix} shows that confusion matrix of previous method at incremental step 4. It clearly shows that most predictions are biased towards the recent entity (class 4). We find that such a bias can be found in the cross-entropy loss function. When learning new entities, the model's gradient update for old entities is defined as:
\begin{equation}
\begin{aligned}
\frac{\partial \mathcal{L}_{ce}}{\partial \omega_{y\prime}} &\propto e^{\Phi_{y\prime}(x_i)}(y^{\prime} \neq y_i),
\end{aligned}
\end{equation}
where the gradient update for old entities is proportional to the classification score for that entity. During the incremental sequence labeling process, this gradient update exhibits an overly penalizing effect on the old entity probability distributions. It shows up as an excessive reduction in the output probability score of the old entity. We provide a more detailed explanation and derivation in Appendix \ref{appendixsec:debiasing in ce}. 

We assume the old model has learned the optimal representation of old entities. Therefore, the new entities should have a smaller impact on the knowledge of old entities. Otherwise, because of the absence of rehearsal samples of the old entities, the model will face catastrophic forgetting of the old entities. In addition, the new entity was not in the predefined set, and a change from a non-entity to a new entity occurs during learning. Therefore, it is reasonable to have a penalizing effect on non-entities, and the debiased cross-entropy loss function is defined as follows:
\begin{equation}
\mathcal{L}^{Debias}_{ce} = \frac{1}{|\mathcal{D}_t|}\sum_{i=1}^{|\mathcal{D}_t|} \log [1+\sum_{y\prime\neq y_i}e^{\mathbf{\delta}\Phi_{y\prime}(x_i)-\Phi_{y_i}(x_i)}],
\label{equ:debias_gradient}
\end{equation}
where $\delta$ is the correction factor for the gradient update of the old entity weights (excluding non-entity weights), $\delta\in[0, 1]$. When $\delta\rightarrow0$, the model will no longer penalize the learning of old entities. When $\delta\rightarrow1$, Eq.\ref{equ:debias_gradient} degenerates to the traditional cross-entropy loss function.
\subsubsection{Learning with Prototypes}
In Section \ref{sec:two shifts}, we elucidate the reasons behind the emergence of O2E at the feature optimization level. In this section, we introduce the utilization of class centers of old entities as class prototypes during the learning process of new entities. Following each task training, we compute prototypes using feature representations from the training set and store them. These prototypes then participate in training the model classifier for the subsequent task alongside the feature representations of new entities.

The class prototypes of old entities serve two essential purposes: firstly, they participate in optimization alongside new entities in each batch, ensuring a balanced optimization process among entities. Secondly, these class prototypes act as anchors in the feature space, mitigating the issue of over-labeling new entities. As depicted in Figure \ref{fig:model}, the introduction of old prototypes reduces the potential over-labeling of new entities, enhancing the precision of new entity learning.

To this end, we defined the loss function of prototypes as follows:
\begin{equation}
\mathcal{L}_{pro} = -\sum_{i=1}^{t-1} \widetilde{y_i} \log \left(\frac{e^{\omega^T \mathcal{P}_i}}{\sum_{j=0}^{\left|\mathcal{Y}_t \right|} e^{\omega^T \mathcal{P}_j}}\right),
\label{equ:Loss_pro}
\end{equation}
where $\widetilde{y_i}$ stands for the label of the old prototype and 
$\mathcal{P}_i$, $\mathcal{P}_j$ stand for old prototypes, defined as follows:
\begin{equation}
\mathcal{P}_t = \frac{1}{|\mathcal{D}_t|}\sum_{i=1}^{|\mathcal{D}_t|} f_{\theta}(x_i).
\label{equ:protptype}
\end{equation}

Our approach differs from OCILNER's approach, which uses prototypes in two ways:(1) OCILNER's approach stores old samples for calculating prototypes. However, in this paper, we only use the training data in each incremental step for calculating prototypes and do not introduce replay samples. (2) OCILNER uses prototypes to label new datasets and adopts a cosine similarity as the threshold for entity labeling. However, in this paper, we found that some of the real non-entities also have a high cosine similarity with entities, which can easily produce wrong labeling for real non-entities and exacerbate semantic migration from entities to non-entities.

In summary, the objective function of our method is defined as follows:
\begin{equation}
\mathcal{L} = \underbrace{{\mathcal{L}_{ce}^{Debias}} + \alpha \mathcal{L}_{pro}}_{\mathcal{L}_{O2E}} + \underbrace{\beta \mathcal{L}_{kd}}_{\mathcal{L}_{E2O}}.
\label{equ:Loss_total}
\end{equation}

\begin{table*}[t]
\caption{Comparisons with state-of-the-art methods on the i2b2 dataset using the \textit{bert-base-cased} model. The best results are highlighted in \textbf{bold} and the second best results are \underline{underlined}. The average of each incremental step is provided in Figure \ref{fig:line_graph}.}
\label{tab:basic_exp_i2b2}
    \centering
\resizebox{\textwidth}{!}{
    \begin{tabular}{c|cc|cc|cc|cc}
    \toprule
          \multirow{2}{*}{Methods}  & \multicolumn{2}{c}{FG-1-PG-1} \vline&  \multicolumn{2}{c}{FG-2-PG-2}   \vline & \multicolumn{2}{c}{FG-8-PG-1} \vline& \multicolumn{2}{c}{FG-8-PG-2}  \\
        \cmidrule(r){2-9}
        & $\mathcal{A}_T$ & $\bar{\mathcal{A}}$ & $\mathcal{A}_T$ & $\bar{\mathcal{A}}$ & $\mathcal{A}_T$ & $\bar{\mathcal{A}}$ & $\mathcal{A}_T$ & $\bar{\mathcal{A}}$ \\ 
         \midrule
         FT   & 2.16\tiny{ ± 0.18} & 14.98\tiny{ ± 0.47} & 7.38\tiny{ ± 1.10} & 25.00\tiny{ ± 0.74} & 2.41\tiny{ ±0.17} & 16.14\tiny{ ± 1.81} & 6.38\tiny{ ± 1.23} & 25.82\tiny{ ± 1.36} \\
           SelfTrain & 17.76\tiny{ ± 1.75} & 37.32\tiny{ ± 2.28} & 36.63\tiny{ ± 6.27} & 54.07\tiny{ ± 3.12} & 7.01\tiny{ ± 3.51} & 27.27\tiny{ ± 3.47} & 24.05\tiny{ ± 6.61} & 47.81\tiny{ ± 2.81} \\
           ExtendNER & 19.54\tiny{ ± 1.59} & 39.10\tiny{ ± 3.17} & 29.20\tiny{ ± 5.86} & 48.26\tiny{ ± 4.05} & 7.83\tiny{ ± 1.42} & 29.03\tiny{ ± 1.15} & 24.00\tiny{ ± 6.40} & 42.53\tiny{ ± 2.92} \\
           CFNER & 34.15\tiny{ ± 4.79} & \underline{50.15\tiny{ ± 2.18}} & \underline{47.21\tiny{ ± 2.99}} & 58.03\tiny{ ± 2.28} & 21.50\tiny{ ± 1.49} & 38.53\tiny{ ± 1.01} & 23.91\tiny{ ± 3.91} & 46.31\tiny{ ± 3.39} \\
           DLD   & 23.03\tiny{ ± 4.08} & 42.87\tiny{ ± 4.35} & 41.05\tiny{ ± 2.79} & 57.28\tiny{ ± 1.37} & 13.10\tiny{ ± 3.05} & 35.12\tiny{ ± 2.24} & 32.01\tiny{ ± 4.47} & 51.66\tiny{ ± 1.71} \\
           RDP   & 28.05\tiny{ ± 1.85} & 47.61\tiny{ ± 2.03} & 44.53\tiny{ ± 2.79} & \underline{59.75\tiny{ ± 1.25}} & 26.83\tiny{ ± 3.01} & 42.02\tiny{ ± 1.57} & 41.43\tiny{ ± 5.32} & \underline{56.92\tiny{ ± 4.07}} \\
           OCILNER & 9.30\tiny{ ± 1.79} & 27.75\tiny{ ± 2.82} & 18.45\tiny{ ± 3.18} & 42.43\tiny{ ± 1.90} & 19.76\tiny{ ± 3.56} & 41.01\tiny{ ± 2.77} & 24.86\tiny{ ± 2.12} & 46.75\tiny{ ± 2.14} \\
           ICE\_PLO & 35.45\tiny{ ± 0.91} & 45.65\tiny{ ± 1.32} & 40.32\tiny{ ± 0.58} & 50.25\tiny{ ± 0.93} & 44.79\tiny{ ± 0.93} & 50.61\tiny{ ± 0.72} & 44.23\tiny{ ± 2.22} & 51.05\tiny{ ± 1.83} \\
           ICE\_O & \underline{36.96\tiny{ ± 1.17}} & 46.93\tiny{ ± 1.07} & 43.29\tiny{ ± 1.79} & 51.24\tiny{ ± 1.70} & \underline{46.24\tiny{ ± 1.36}} & \underline{51.70\tiny{ ± 0.85}} & \underline{49.10\tiny{ ± 1.33}} & 53.56\tiny{ ± 1.22} \\
           CPFD  & 17.72\tiny{ ± 3.95} & 46.11\tiny{ ± 1.45} & 31.44\tiny{ ± 5.19} & 53.84\tiny{ ± 2.39} & 5.0\tiny{ ± 3.97} & 32.86\tiny{ ± 3.49} & 23.03\tiny{ ± 7.47} & 50.26\tiny{ ± 3.38} \\
    
        \midrule
        \rowcolor{black!10} \textbf{IS3 (Ours)}  & \textbf{43.88\tiny{ ± 2.05}} & \textbf{56.87\tiny{ ± 0.56}} & \textbf{54.84\tiny{ ± 1.35}} & \textbf{61.83\tiny{ ± 0.87}} & \textbf{50.75\tiny{ ± 1.28}} & \textbf{58.38\tiny{ ± 1.35}} & \textbf{56.96\tiny{ ± 0.68}} & \textbf{63.03\tiny{ ± 1.07}} \\	 
          \bottomrule
    \end{tabular}}
\end{table*}

\begin{figure*}[!t]
 	\begin{subfigure}{0.24\linewidth}
		\includegraphics[width=1\linewidth]{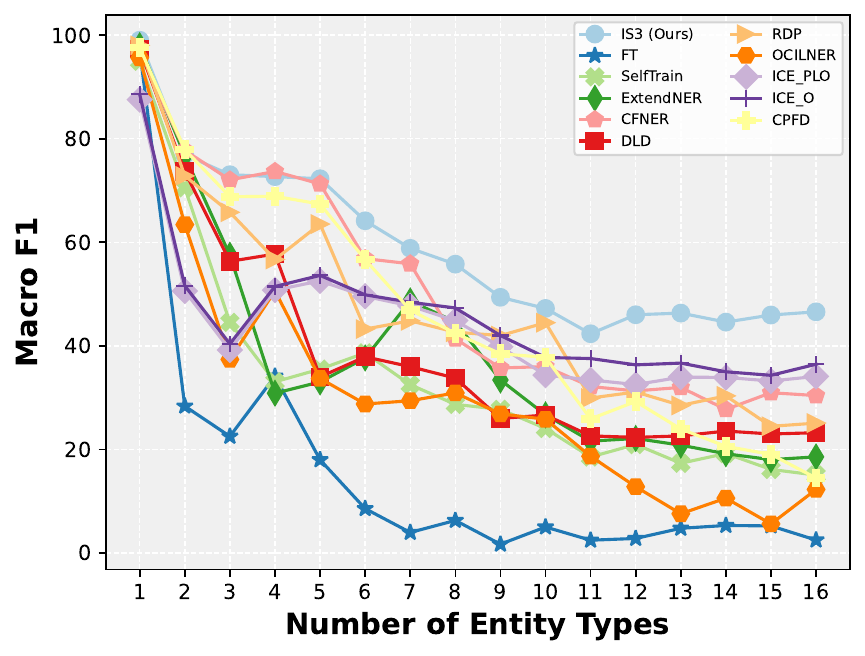}
        \caption{i2b2 (FG-1-PG-1)}
        \label{fig:i2b2a}
	\end{subfigure}
	\begin{subfigure}{0.24\linewidth}
		\includegraphics[width=1\linewidth]{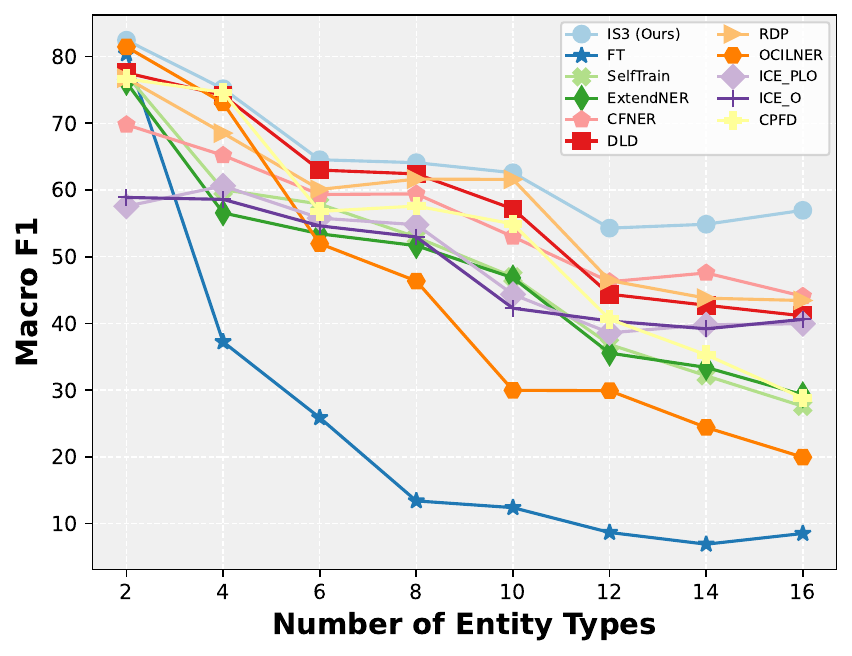}
        \caption{i2b2 (FG-2-PG-2)}
        \label{fig:i2b2b}
	\end{subfigure}
        \begin{subfigure}{0.24\linewidth}
		\includegraphics[width=1\linewidth]{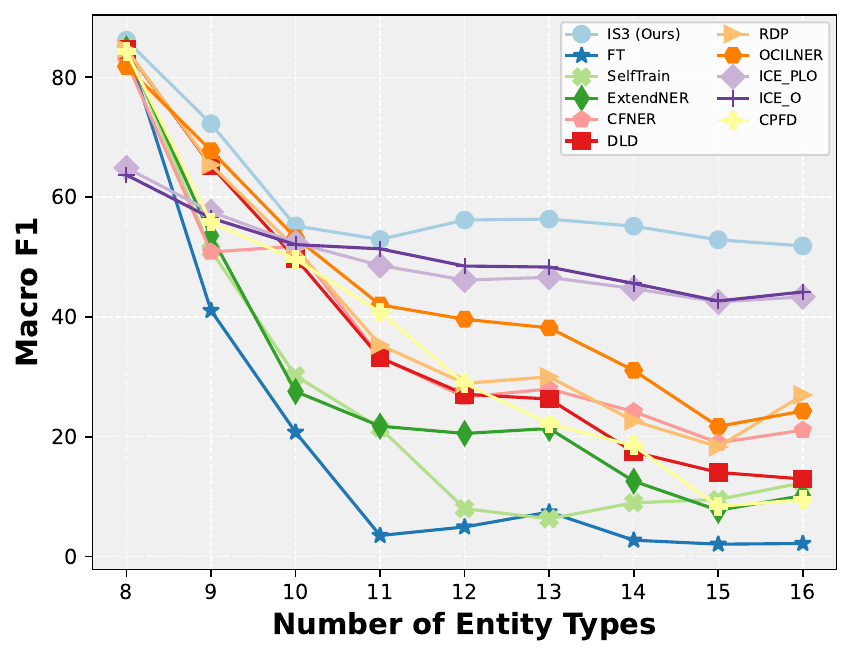}
        \caption{i2b2 (FG-8-PG-1)}
        \label{fig:i2b2c}
	\end{subfigure}
        \begin{subfigure}{0.24\linewidth}
		\includegraphics[width=1\linewidth]{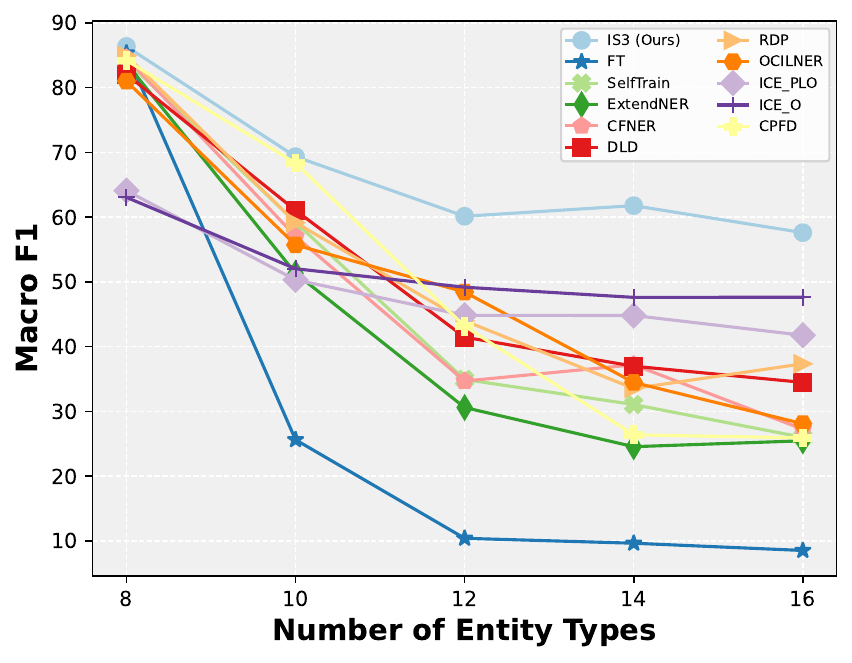}
        \caption{i2b2 (FG-8-PG-2)}
        \label{fig:i2b2d}
	\end{subfigure}
 
        \begin{subfigure}{0.24\linewidth}
		\includegraphics[width=1\linewidth]{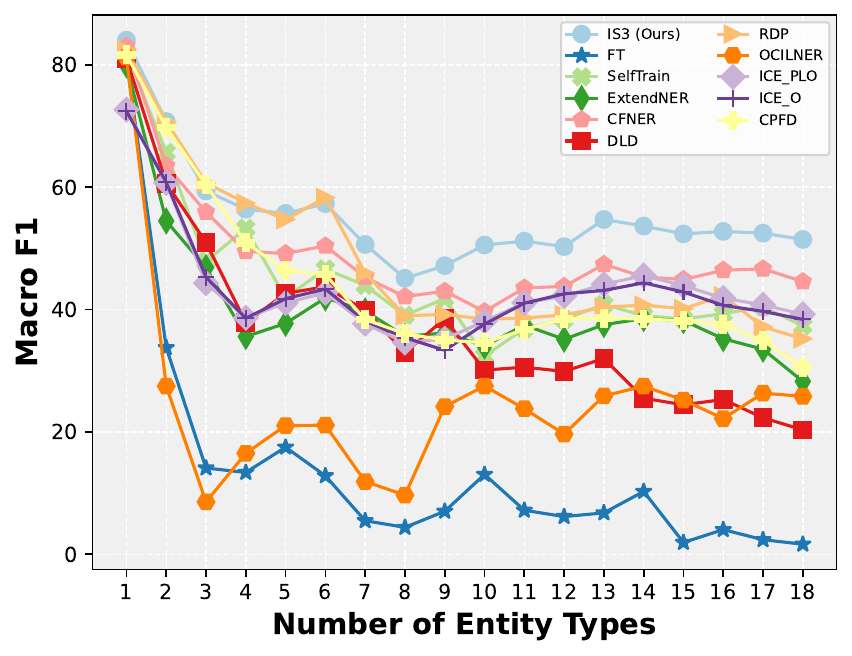}
        \caption{OntoNotes5 (FG-1-PG-1)}
        \label{fig:OntoNotes5a}
	\end{subfigure}
	\begin{subfigure}{0.24\linewidth}
		\includegraphics[width=1\linewidth]{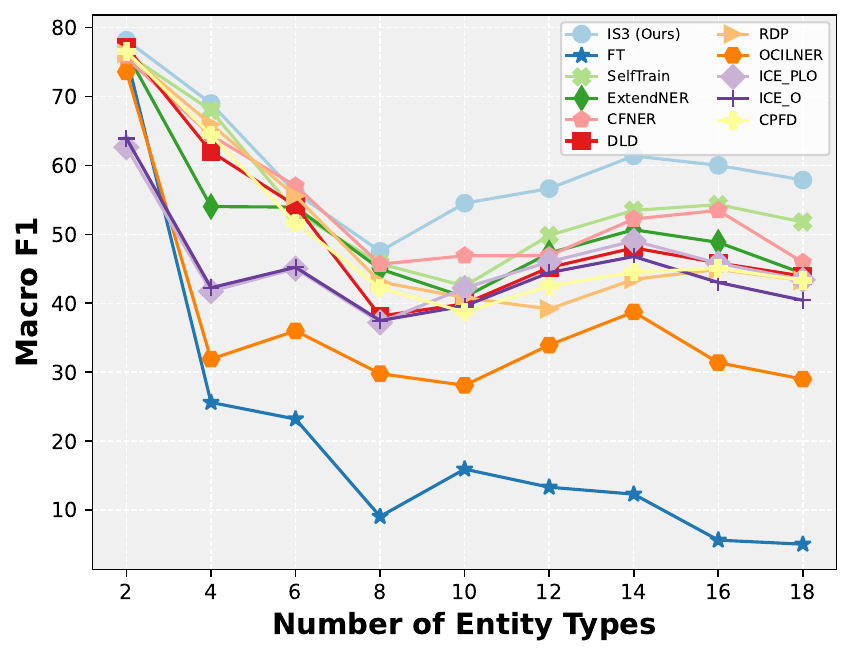}
        \caption{OntoNotes5 (FG-2-PG-2)}
        \label{fig:OntoNotes5b}
	\end{subfigure}
        \begin{subfigure}{0.24\linewidth}
		\includegraphics[width=1\linewidth]{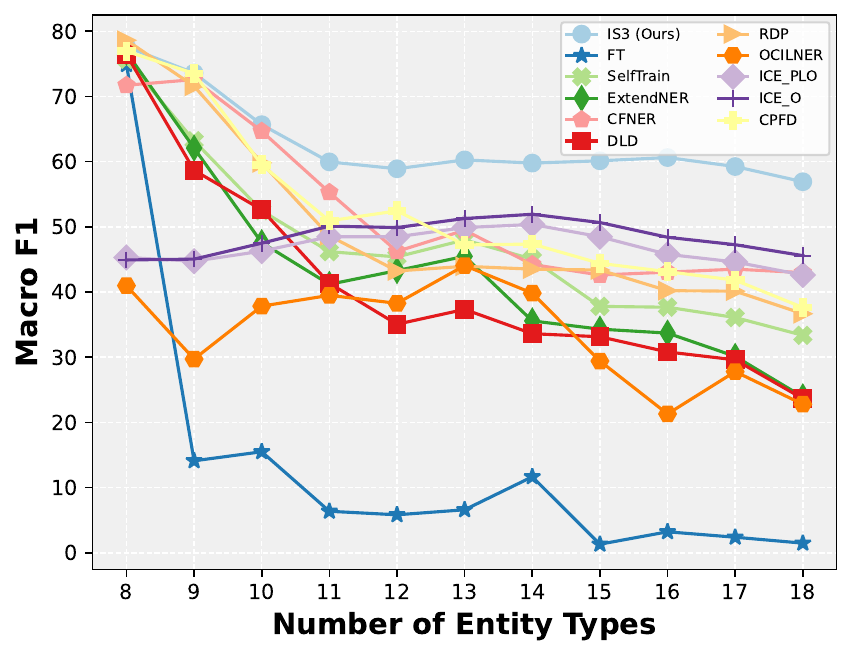}
        \caption{OntoNotes5 (FG-8-PG-1)}
        \label{fig:OntoNotes5c}
	\end{subfigure}
        \begin{subfigure}{0.24\linewidth}
		\includegraphics[width=1\linewidth]{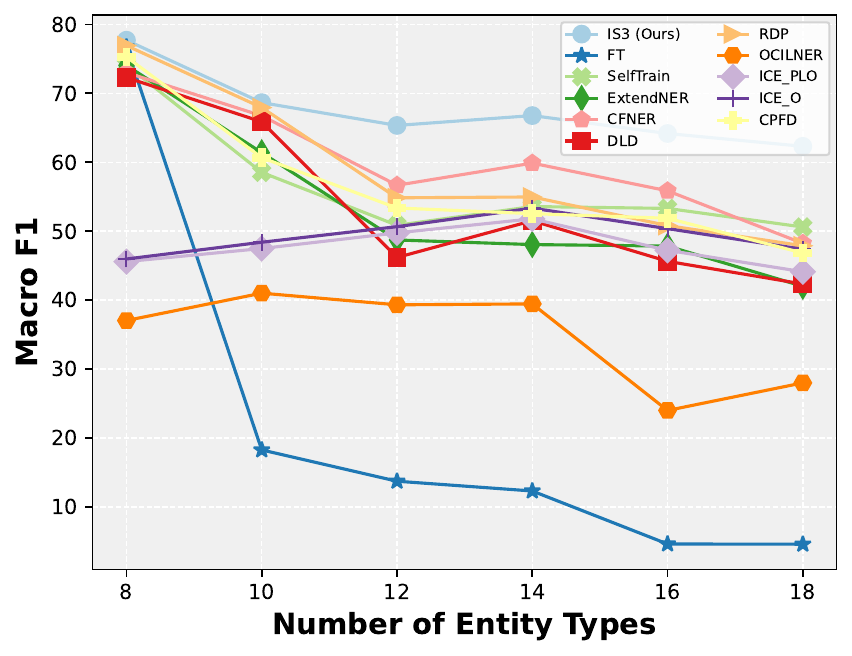}
        \caption{OntoNotes5 (FG-8-PG-2)}
        \label{fig:OntoNotes5d}
	\end{subfigure}
        \caption{Comparison of the step-wise Macro F1 score on i2b2 and OntoNotes5.}
        \label{fig:line_graph}
\end{figure*}

\section{Experiments}
\subsection{Experimental Setup}
\textbf{Datasets}\quad We conducted experiments on three widely used datasets: i2b2 \cite{murphy2010serving}, OntoNotes5 \cite{hovy2006ontonotes}, and MAVEN \cite{wang2020MAVEN}. We divide the dataset into disjoint slices according to categories. In each slice, we keep only the category labels visible to the current task, and the rest of the labels are labeled as non-entities.\\
\textbf{Settings}\quad We sort the above slices according to initial letter and train them in a FG-\textit{a}-PG-\textit{b} manner. FG means that the pre-trained model is trained with \textit{a} entity types as the initial model, and PG means that the initial model is trained with \textit{b} entity types at each following incremental step.\\
\textbf{Baselines}\quad We consider the following state-of-the-art methods for incremental sequence labeling: Self-Training \cite{rosenberg2005semi,de2019continual},
ExtendNER \cite{monaikul2021continual},
CFNER \cite{zheng-etal-2022-distilling},
DLD \cite{zhang2023decomposing},
RDP \cite{zhang2023task},
OCILNER \cite{ma-etal-2023-learning},
ICE \cite{liu-huang-2023-teamwork},
CFPD \cite{zhang2023continual}. Detailed descriptions of the baselines and their experimental setup are provided in Appendix \ref{appendixsec:baselines}.
\begin{figure}[!t]
\centering
\includegraphics[width=0.96\linewidth]{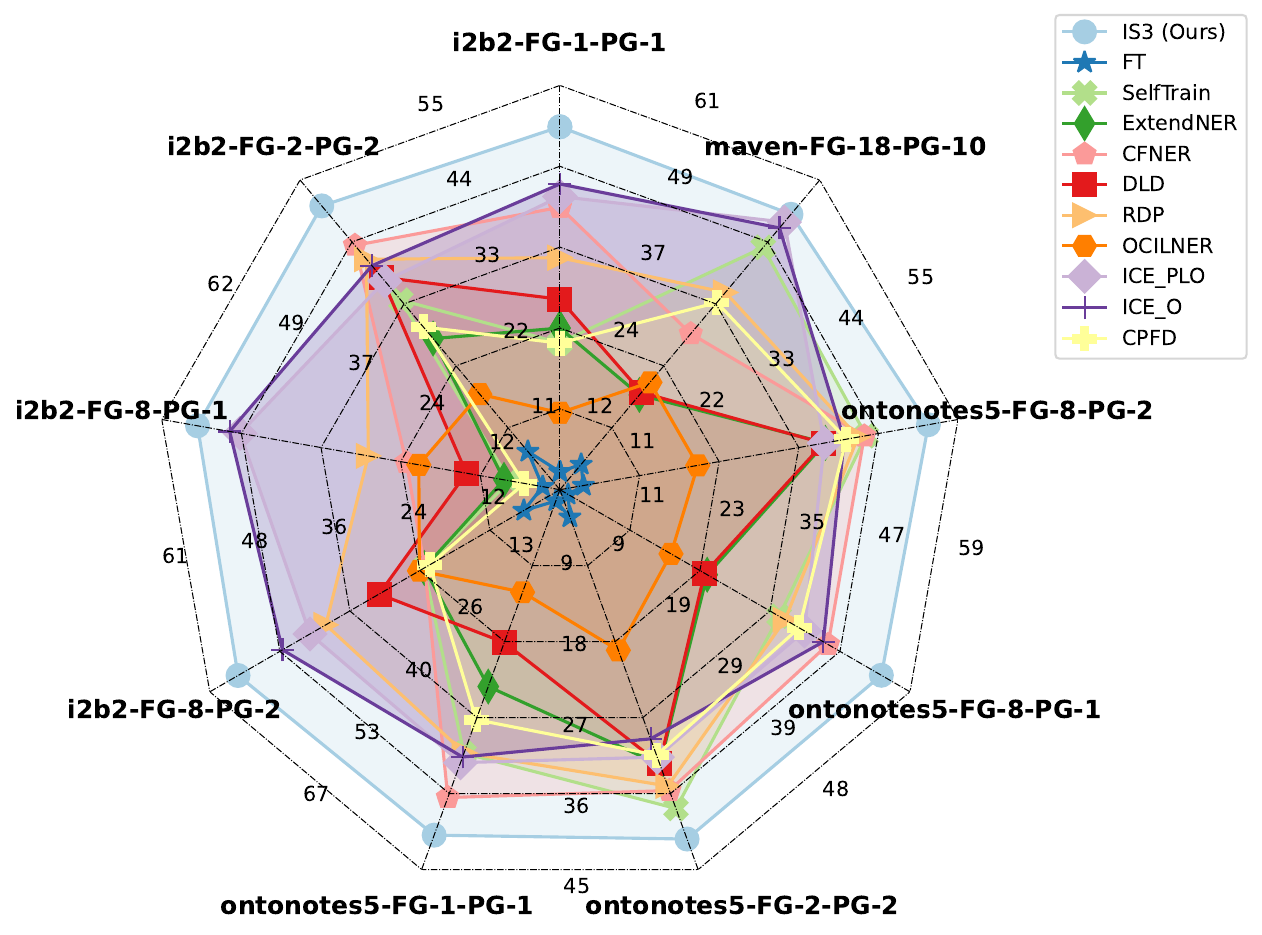}
\caption{The comparison between our method and previous state-of-the-art methods on nine incremental learning settings. We report the MacroF1 score after learning the final task. The detailed results are provided in Table \ref{tab:basic_exp_i2b2}, \ref{tab:basic_exp_i2b2_robert}, \ref{tab:basic_exp_OntoNotes5}, \ref{tab:basic_exp_OntoNotes5_robert}, and \ref{tab:MAVEN_exp}.}
\label{fig:radar}
\end{figure}\\
\textbf{Implementation Details}\quad We use \textit{bert-base-cased} model from HuggingFace \cite{wolf2019huggingface} as backbone, with a hidden dimension of $d=768$. In addition to this, we have included supplemental experiments with the \textit{roberta-base} model on the relevant datasets in Appendix \ref{appendixsec:Additional Experimental Results}. We use the AdamW  \cite{loshchilov2018decoupled} optimizer, with learning rate $1e^{-6}$ and $1e^{-3}$ for backbone and classifier. We report the mean and standard deviation results over five runs.\\
\textbf{Metrics}\quad Considering that each of the categories should have a comparable degree of contribution in the test, we use Macro F1 to evaluate the performance of the model. We use the last step Macro F1 result in $\mathcal{A}_T$, and the average Macro F1 result in $\bar{\mathcal{A}}$, on all incremental steps as evaluation metrics.
$\mathcal{A}_T$ and $\bar{\mathcal{A}}$ are defined in Appendix \ref{appendixsec:Metrics}.
\subsection{Results and Analysis}

\noindent\textbf{Comparisons with State-Of-The-Art}\quad To validate the effectiveness of our approach, we conducted exhaustive experiments on the i2b2, OntoNotes5, and MAVEN datasets. We used the Finetune Only (FT) approach as a lower bound for comparison. Table \ref{tab:basic_exp_i2b2}, \ref{tab:basic_exp_i2b2_robert} displays the results of the experiments conducted on i2b2. Due to space limitations, we provide the results on MAVEN in Table \ref{tab:MAVEN_exp} and OntoNotes5 in Table \ref{tab:basic_exp_OntoNotes5}, \ref{tab:basic_exp_OntoNotes5_robert}. In detail, we show the experimental results under nine incremental learning settings through Figure \ref{fig:radar}. Our method consistently outperforms the previous state-of-the-art method in multiple settings, from FG-1-PG-1 to FG-8-PG-2. The poor performance of the previous method may be attributed to the ignorance of O2E.
\begin{figure}[!t]
	\centering
 	\begin{subfigure}{0.49\linewidth}
		\centering
		\includegraphics[width=1\linewidth]{./fig/extendner_confusion_v3.pdf}
                  \caption{ExtendNER}
        \label{fig:confusion matrixa}

		\label{texttsne}
	\end{subfigure}
	\begin{subfigure}{0.49\linewidth}
		\centering
		\includegraphics[width=1\linewidth]{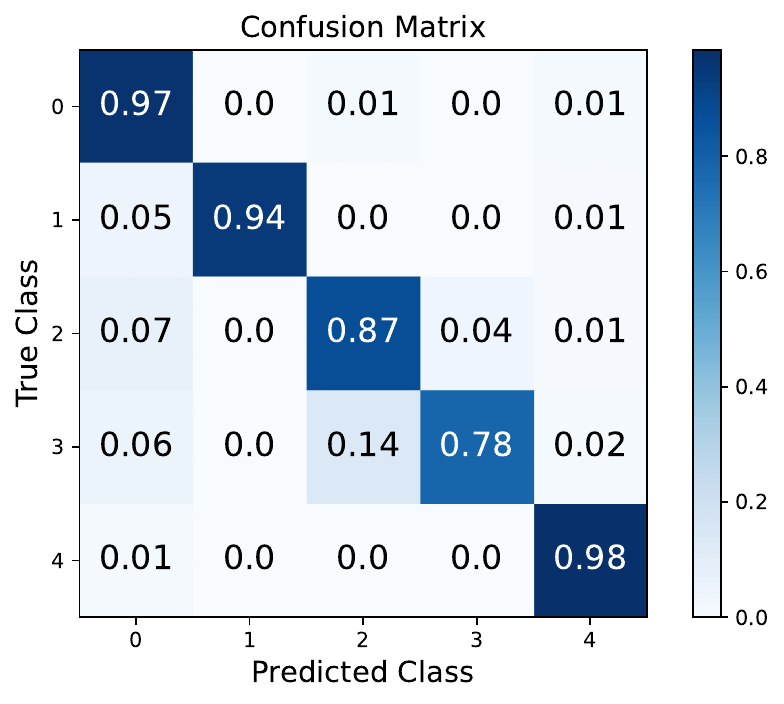}
                \caption{IS3 (Ours)}
        \label{fig:confusion matrixb}
	\end{subfigure}
        \caption{Visualization of prediction of previous method and IS3 approach in task 4. Our approach greatly mitigates the E2O and O2E shift problems and balances the old and new classes well on the model predictions.}
        \label{fig:contrast}
\end{figure}
\definecolor{color1}{RGB}{248, 203, 173}
\definecolor{color2}{RGB}{169, 209, 142}
\begin{figure*}[!t]
\centering
    \includegraphics[width=0.98\linewidth]{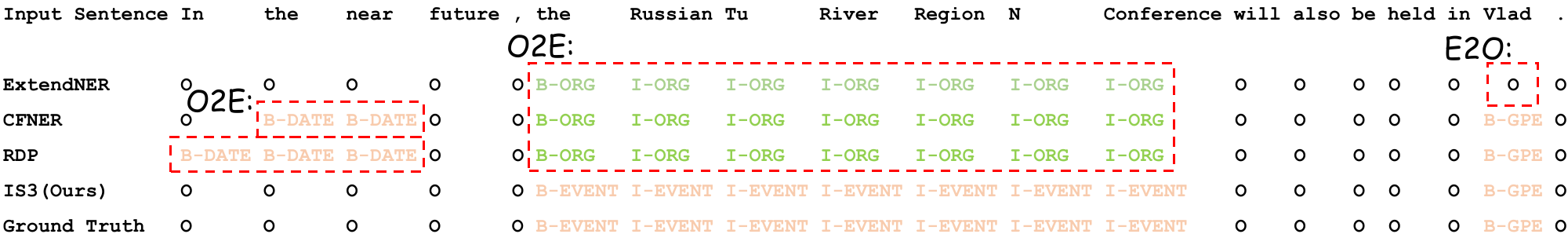}
    \caption{A sample from OntoNotes5. \textcolor{color1}{EVENT, DATE, GPE} are old entites. \textcolor{color2}{ORG} is a new entity. The previous method had the issue of mislabeling non-entities as old entities and overwriting old entities as new ones. In contrast, our method accurately labels old entities when learning the new entity, demonstrating its effectiveness and superiority.}
    \label{fig:case_study}
\end{figure*}
As shown in Figure \ref{fig:contrast}, during the learning process, the previous method, ExtendNER, confuses new entities with non-entities due to O2E and old entities with non-entities due to E2O. Both of them together lead to poor prediction results of the model. We have effectively mitigated the above problems through our framework IS3, which strikes a good balance between maintaining old entities and learning new ones. 

\begin{figure*}[!t]
	\centering
 	\begin{subfigure}{0.40\linewidth}
		\centering
		\includegraphics[width=1\linewidth]{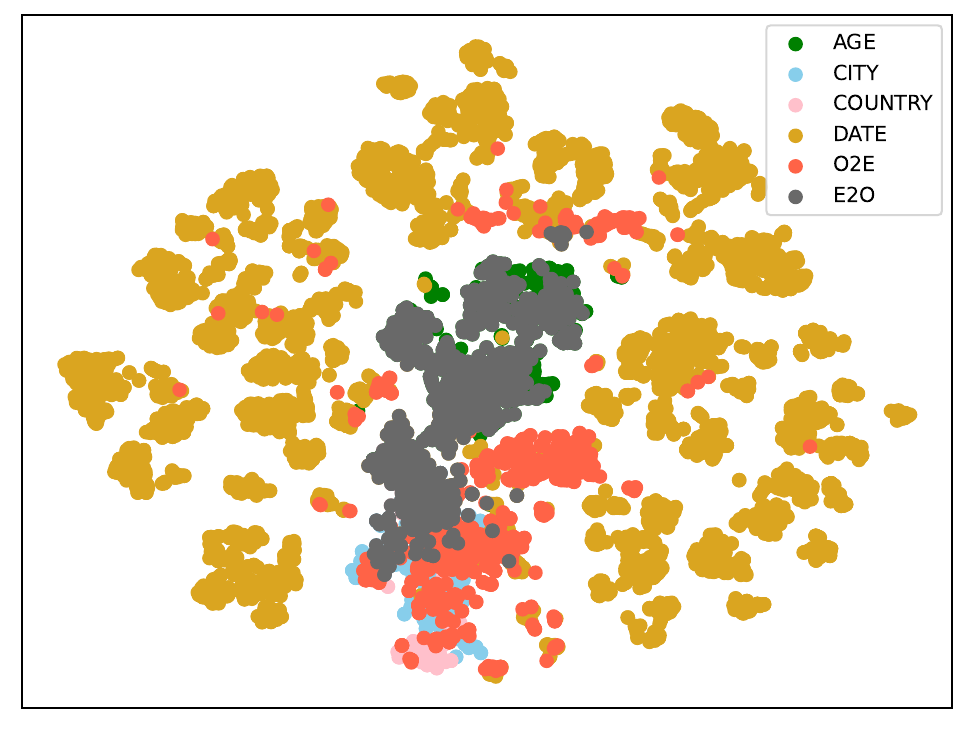}
                  \caption{ExtendNER}
        \label{fig:tsne_extendner}
	\end{subfigure}
        \begin{subfigure}{0.40\linewidth}
		\centering
		\includegraphics[width=1\linewidth]{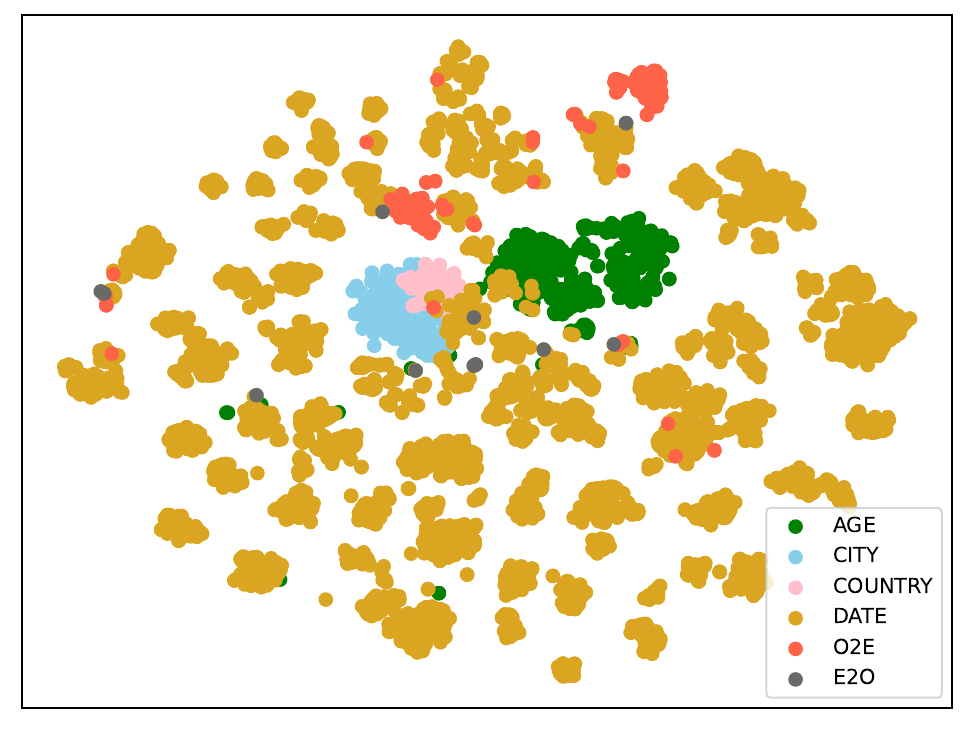}
                \caption{IS3 (Ours)}
        \label{fig:tsneours}
	\end{subfigure}
        \caption{The T-SNE visualization of the feature representations on ExtendNER and our method. Our approach IS3 greatly mitigates O2E and E2O, resulting in good discrimination between old and new entities.}
        \label{fig:tsne}
\end{figure*}

To further demonstrate the effectiveness of our method, we visualize the feature representation through T-SNE \cite{van2008visualizing}. As shown in Figure \ref{fig:tsne}, the ExtendNER method faces serious E2O and O2E problems, with new entities and non-entities overwriting old ones, leading to catastrophic forgetting. Our method successfully addresses the issue of semantic bias that arises when the model learns a new task.

\begin{table}[!t]
\caption{The ablation study of our method on i2b2 and OntoNotes5 under the setting FG-1-PG-1, MAVEN under the setting FG-18-PG-10. The ablation of each component resulted in a significant decrease in model performance, proving the effectiveness of all our components.}
\label{tab:ablation exp}
\centering
\resizebox{\linewidth}{!}{
    \begin{tabular}{c|cc|cc|cc}
    \toprule
         \multirow{2}{*}{Methods}  & \multicolumn{2}{c}{i2b2} \vline&  \multicolumn{2}{c}{OntoNotes5}   \vline & \multicolumn{2}{c}{MAVEN}\\
        \cmidrule(r){2-7}
        & $\mathcal{A}_T$ & $\bar{\mathcal{A}}$ & $\mathcal{A}_T$ & $\bar{\mathcal{A}}$ & $\mathcal{A}_T$ & $\bar{\mathcal{A}}$  \\ 
         \midrule
         \rowcolor{black!10} IS3 (Ours)   & 43.88\tiny{ ± 2.05} & 56.87\tiny{ ± 0.56} & 50.23\tiny{ ± 0.94} & 54.65\tiny{ ± 0.84} & 40.15\tiny{ ± 0.38} & 48.16\tiny{ ± 0.16}  \\
           w/o $\mathcal{L}^{Debias}_{ce}$ & 40.79\tiny{ ± 0.89} & 54.39\tiny{ ± 0.19} & 47.89\tiny{ ± 0.91} & 52.77\tiny{ ± 1.21} & 38.19\tiny{ ± 0.98} & 46.56\tiny{ ± 0.58}  \\
           w/o $\mathcal{L}_{pro}$ & 25.88\tiny{ ± 2.78} & 45.95\tiny{ ± 2.53} & 44.26\tiny{ ± 1.33} & 50.07\tiny{ ± 1.08} & 34.64\tiny{ ± 0.78} & 45.15\tiny{ ± 0.39}  \\
           w/o Both  & {23.22\tiny{ ± 2.12}} & {37.81\tiny{ ± 3.81}} & {42.77\tiny{ ± 0.22}} & {49.11\tiny{ ± 0.49}} & {31.03\tiny{ ± 0.34}} & {42.61\tiny{ ± 0.87}} \\
          \bottomrule
    \end{tabular}}
\end{table}

\noindent\textbf{Ablation Study}\quad We explored the validity of the components of our approach through ablation experiments, and the results are shown in Table \ref{tab:ablation exp}. We removed the debiased cross-entropy loss $\mathcal{L}^{Debias}_{ce}$ and prototype loss $\mathcal{L}_{pro}$ modules, respectively. These results demonstrate the essential roles played by both $\mathcal{L}^{Debias}_{ce}$ and $\mathcal{L}_{pro}$ modules. The $\mathcal{L}^{Debias}_{ce}$ reduces the penalizing effect of the new entity on the old entity and enhances the discrimination between the old and new entities by improving the prediction confidence of the old entity. The $\mathcal{L}_{pro}$ corrects the bias of modeling new entities by shrinking the scope of over-labeling new entities through old prototypes.

\begin{figure}[!t]
\centering
    \includegraphics[width=0.96\linewidth]{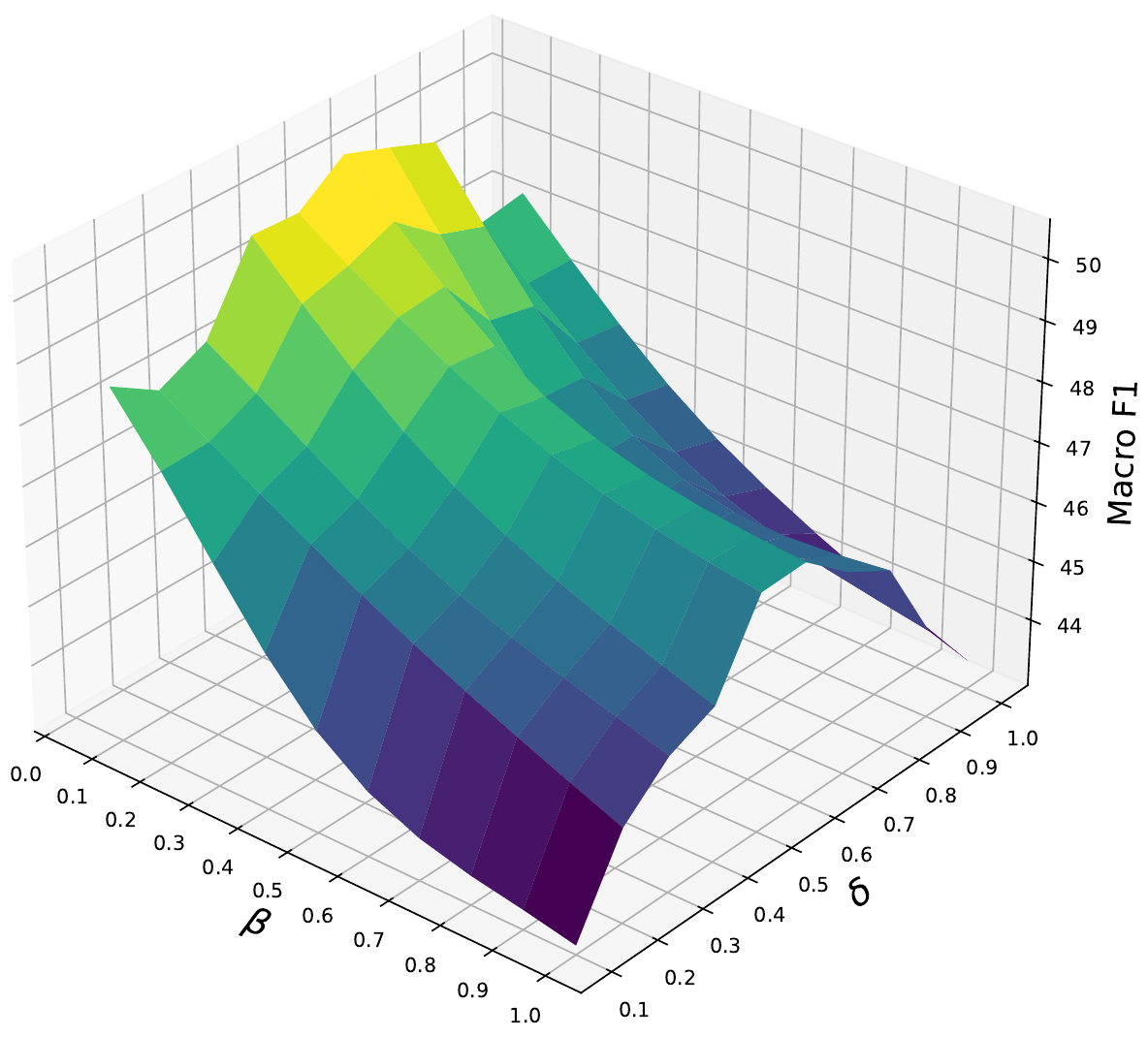}
    \caption{The results of different hyper-parameter choices on i2b2 with the setting FG-1-PG-1. We show the results are $\delta \in(0,1]$ and $\beta \in(0,1]$. }
    \label{fig:hyperparameter}
\end{figure}

\noindent\textbf{Hyper-Parameter Analysis}\quad Figure \ref{fig:hyperparameter} shows the results of different hyper-parameter choices on OntoNotes5 with the setting FG-1-PG-1. We consider two hyper-parameters: the correction factor in the debiased cross-entropy loss $\delta$ and the weight of the prototype loss $\beta$. The results show that $\delta$ around 0.5 reaches the best result, indicating that a moderate penalty effect reduction favors model performance. As $\beta$ keeps increasing, it makes the model overfit for the old prototype, leading to a decrease in model performance.

\noindent\textbf{Case Study}\quad We provide an example in Figure \ref{fig:case_study} to demonstrate that the previous method suffers from an O2E offset when learning a new entity ORG, overwriting the old entity EVENT as a new entity. Simultaneously, the model inherits past O2E issues (labeling [O] as [DATE]). Additionally, it suffers from E2O, which fails to recognize the old entity accurately. Our method effectively balances these two offset problems and is more conducive to model learning.
\section{Conclusion}
In this paper, we introduce a novel perspective on the catastrophic forgetting problem in incremental sequence annotation, identifying and addressing both E2O and O2E semantic shifts. Bridging gaps in previous research, we propose the IS3 framework to tackle both issues. Comprehensive experiments on three datasets demonstrate that our IS3 method significantly outperforms previous state-of-the-art approaches. This work provides a fresh outlook on the incremental sequence labeling task and offers effective solutions to mitigate the catastrophic forgetting problem.

\section*{Limitations}
While the proposed method effectively mitigates catastrophic forgetting to some extent, its reliance on the predictions of old models for preserving existing knowledge can result in accumulated prediction errors, which may lead to poor model performance in more incremental steps. Moreover, the current method does not thoroughly explore the relationship between the penalty effect and the dataset, leaving potential avenues for future research.

\section*{Acknowledgements}
The work described in this paper was partially funded by the National Natural Science Foundation of China (Grant No. 62272173), the Natural Science Foundation of Guangdong Province (Grant Nos. 2024A1515010089, 2022A1515010179), the Science and Technology Planning Project of Guangdong Province (Grant No. 2023A0505050106), and the National Key R\&D Program of China (Grant No. 2023YFA1011601).

\bibliography{anthology,custom}

\clearpage
\appendix
\section{Derivation of Debiased Cross-entropy Loss Function}
\label{appendixsec:debiasing in ce}
The overall model parameters are defined as $\Theta = \{\theta, \omega\}$. The model's backbone $f_\theta:X\rightarrow\mathbb{R}^d$ extracts feature embeddings of dimension $d$ from the inputs. Following the backbone, a linear classifier produces logits $\Phi(\cdot) = \omega^T\cdot f_\theta(\cdot):X\rightarrow\mathbb{R}^{\left| \mathcal{Y}_t \right|}$, where $\omega$ represents the classifier weights for the corresponding dimensions and $\mathcal{Y}_t$ represents the current label set. The softmax probability of new entity is defined as: $p_{y_i} = \frac{e^{\Phi_{y_i} (x_i)}}{\sum_{y^\prime\in \mathcal{Y}_t}e^{\Phi_{y^\prime}}(x_i)}$. The derivation of Debiased Cross-entropy Loss Function is proved as follows:
\begin{equation}
\begin{aligned}
\frac{\partial \mathcal{L}_{ce}}{\partial \omega_{y\prime}} &= \frac{\partial \mathcal{L}_{ce}}{\partial p_{y_i}} \cdot {\frac{\partial p_{y_i}}{\partial {\Phi_{y^\prime}}(x_i)}} \cdot \frac{\partial {\Phi_{y^\prime}}(x_i)}{\partial \omega_{y\prime}} \\
&= -\frac1{\ln{2}\cdot p_{y_i}} \cdot f_\theta(x_i) \cdot {\frac{\partial p_{y_i}}{\partial {\Phi_{y^\prime}}(x_i)}}\\
&= \frac{f_\theta(x_i)}{\ln{2}} \cdot {p_{y^\prime}} \propto e^{\Phi_{y\prime}(x_i)}(y^{\prime} \neq y_i)
\end{aligned}
\end{equation}

for the same input $x_i$, $\frac{f_\theta(x_i)}{\ln{2}}$ can be viewed as a constant. Therefore, the gradient penalty of the new entity over the old entity is proportional to the probability value of the old entity.

\section{Datasets}
\label{appendixsec:datasets}

Table \ref{tab:dataset_introduction} shows the detailed description of each dataset. Table \ref{tab:dataset_format} shows the format of inputs and outputs for Sequence labeling task.
\textbf{FL} means the \textit{full ground-truth label} for all steps. During the learning process, we will label unseen entities as non-entities [O].

\section{Baselines}
\label{appendixsec:baselines}
The introduction about the baselines in the experiment and their settings are as follows:
\begin{itemize}
    \item SelfTrain \cite{rosenberg2005semi,de2019continual}: SelfTrain utilizes the labels generated by the predictions of the old model on the new dataset, combined with the labels of the new entities, to guide the training of the new model.
    \item ExtendNER \cite{monaikul2021continual}: ExtendNER introduces knowledge distillation to review the knowledge of old entities, aiming to align the outputs of the old and new models for old entities using KL divergence. In contrast to SelfTrain, ExtendNER retains specific structural information through the probability distribution of the model output. The coefficient of the distillation loss $\lambda = 2$.
    \item CFNER \cite{zheng-etal-2022-distilling}: CFNER proposes a unified causal framework to extract causality from both new entity types and the Other-Class and employs curriculum learning to alleviate the impact of label noise and introduce a self-adaptive weight to balance the causal effects between new entity types and the Other-Class. The number of matched tokens $K = 3$, the initial value of balancing weight $\lambda_{base} = 2$ and the initial value of confidence threshold $\delta_1 = 1$.
    \item DLD \cite{zhang2023decomposing}: DLD decomposes a prediction logit into two terms, measuring the probability of an input token belonging to a specific entity type or not. The coefficient of the distillation loss $\lambda = 2$.

\begin{table*}[!t]
\caption{Detailed description of each dataset.}
\label{tab:dataset_introduction}
\centering
\resizebox{0.95\linewidth}{!}{
    \begin{tabular}{cccl}
        \toprule
         \textbf{Dataset}& \textbf{Entity Type} &\textbf{Sample}& \textbf{Entity Type Sequence (Alphabetical Order)}\\
         \midrule
         &&& AGE, CITY, COUNTRY, DATE, DOCTOR, HOSPITAL,\\
         \multirow{2}{*}{i2b2} & \multirow{2}{*}{16} & \multirow{2}{*}{141k}
         & IDNUM, MEDICALRECORD, ORGANIZATION, \\
         &&& PATIENT, PHONE, PROFESSION, STATE, STREET, \\
         &&& USERNAME, ZIP\\
         \midrule
         &&& CARDINAL, DATE, EVENT, FAC, GPE, LANGUAGE,\\
         \multirow{2}{*}{OntoNotes5}&\multirow{2}{*}{18} &\multirow{2}{*}{77k}
         & LAW, LOC, MONEY, NORP, ORDINAL, ORG,\\
         &&& PERCENT, PERSON, PRODUCT, QUANTITY, TIME,\\
         &&& WORK\_OF\_ART\\
         \midrule
         \multirow{2}{*}{MAVEN}&\multirow{2}{*}{178}&\multirow{2}{*}{124k}&ACTION, ARREST, BRINGING, CONTROL, EXPANSION, \\
         &&&INCIDENT, INFLUENCE, VIOLENCE etc.\\
         \bottomrule
    \end{tabular}}
\end{table*}

\begin{table*}[!t]
\caption{Examples of inputs and outputs for each dataset.}
\label{tab:dataset_format}
\centering
\resizebox{0.95\linewidth}{!}{
    \begin{tabular}{cl}
        \toprule
         \textbf{Inputs} & Xinhua\quad news\quad agency , \ Beijing \thinspace, \enspace August \quad 31st \\
         \textbf{FL} & B-ORG \thinspace I-ORG \ I-ORG O B-GPE O B-DATE I-DATE\\
         \midrule
         \textbf{Inputs} &There were no direct\enspace effects\quad of the earthquake\quad \enspace '\enspace s\\
         \midrule
         \textbf{FL} & O\qquad O\quad\ O\quad O B-Influence O\enspace O B-Catastrophe O O\\
         & shaking\enspace due to its low intensity.\\
         &  B-Motion O\enspace O\enspace O\enspace O\quad O\\
         \bottomrule
    \end{tabular}}
\end{table*}
    \item RDP \cite{zhang2023task}: RDP introduces a task relation distillation scheme with two aims: ensuring inter-task semantic consistency by minimizing inter-task relation distillation loss and enhancing model prediction confidence by minimizing intra-task self-entropy loss. The coefficient of inter-task relation distillation loss $\lambda_1 = 0.3$ and the coefficient of intra-task self-entropy loss $\lambda_2 = 0.1$.
    \item OCILNER \cite{ma-etal-2023-learning}: OCILNER introduces a novel representation learning method aimed at acquiring discriminative representations for entities and non-entities, which can dynamically identify entity clusters within non-entities. The threshold for relabeling samples $\beta_i = 0.98 - 0.05 * (t-i)$, where t is the current step, and i is the id of the old task. 
    \item ICE \cite{liu-huang-2023-teamwork}: ICE freezes the backbone model and the old entity classifiers, focusing solely on training new entity classifiers. This approach includes two methods: ICE\_O and ICE\_PLO. The former combines logits of non-entity with logits of new entities for output probability computation during training, while the latter combines all previous logits with new entity logits.
    \item CFPD \cite{zhang2023continual}: CPFD introduces a pooled feature distillation loss that adeptly balances the trade-off between retaining knowledge of old entity types and acquiring new ones and a confidence-based pseudo-labeling method for the non-entity type. The balancing weight $\lambda = 2$.
\end{itemize}

\section{Metrics}
\label{appendixsec:Metrics}
The last step Macro F1 result $\mathcal{A}_T$ and the average Macro F1 result $\bar{\mathcal{A}}$ are defined as follows:
\begin{equation}
a_{t}=\frac{1}{|\mathcal{D}_t|}\sum_{i=1}^{|\mathcal{D}_t|}\mathbb{1} (\operatorname*{argmax}_{y^{\prime}\in\mathcal{Y}_i}\Phi_{t, y^{\prime}}(x_i)=y_i), 
\label{equ:acc}
\end{equation}
where $a_{t}$ represents the F1 score of the $t^{th}$ entity, $|\mathcal{D}_t|$ repesents the number of entities and $\mathbb{1}(\cdot)$ is the indicator for $\Phi_{t, y^{\prime}}(x_i)=y_i$. 

\begin{equation}
\mathcal{A}_T = \frac{1}{N}\sum_{j=1}^{N} a_{j},
\label{equ:marcro_f1}
\end{equation}
where $\mathcal{A}_T$ stands for the MacroF1 score at incremental step $t$.

\begin{equation}
\bar{\mathcal{A}} = \frac{1}{N}\sum_{k=1}^{N} {A}_{T_k},
\label{equ:average_marcro_f1}
\end{equation}
where $\bar{\mathcal{A}}$ stands for the average MacroF1 score for all incremental steps.

\begin{figure}[htbp]
\centering
    \includegraphics[width=0.98\linewidth]{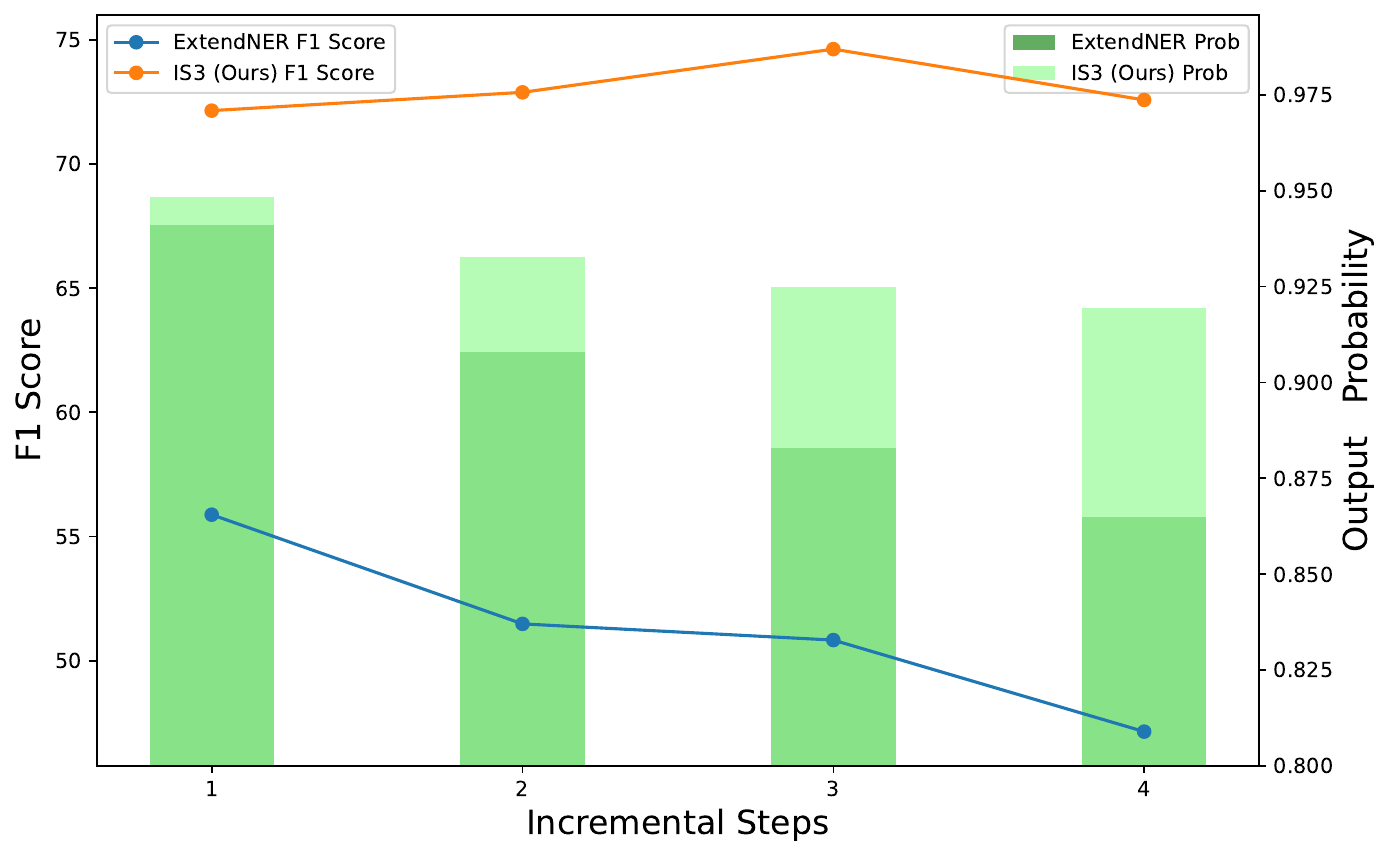}
    \caption{The F1 score and probability distributions of class "DATE" in OntoNotes5 with incremental steps. }
    \label{fig:logits_comparison}
\end{figure}

\section{Additional Experimental Results}
\label{appendixsec:Additional Experimental Results}
Table \ref{tab:MAVEN_exp} highlights the superiority of our method IS3 over previous methods on MAVEN. However, due to its larger number of classes, the performance of the model decreases in subsequent incremental steps. Table \ref{tab:basic_exp_OntoNotes5} shows the results of the experiments conducted on OntoNotes5. Our method IS3 achieves improvements over
the previous SOTA ranging from 5.47\% to 10.52\% in MarcoF1 score, and 3.89\% to 6.53\%, under four settings (FG-1-PG-1, FG-2-PG-2, FG-8-PG-1, and FG-8-PG-
2) of the OntoNotes5 dataset. 

As shown in Figure \ref{fig:logits_comparison}, the previous method exhibits a rapid decrease in probability distribution with increasing incremental steps, coinciding with a decline in the F1 score. In contrast, our approach IS3 effectively mitigates the model's penalization of old entities, thereby maintaining good performance.

\begin{table}[htbp]
  \centering
  \caption{The comparison of training time and trainable parameters for each task on OntoNotes5.}
  \resizebox{\linewidth}{!}{
        \begin{tabular}{ccc}
    \toprule
          & \textbf{\# Time (Min)} & \textbf{\# Trainable Params each Task} \\
    \midrule
    SEQ   & 150   & 109M \\
    SelfTrain & 276   & 109M \\
    ExtendNER & 182   & 109M \\
    CFNER & 512   & 109M \\
    DLD   & 158   & 109M \\
    RDP   & 188   & 109M \\
    OCILNER & 420   & 109M \\
    ICE   & 126   & 28K \\
    CPFD  & 282   & 109M \\
    \midrule
    IS3(Ours)  & 155   & 109M \\
    \bottomrule
    \end{tabular}%
     }
  \label{tab:comparison_time_params}%
\end{table}%

\begin{table}[htbp]
\caption{Comparisons with state-of-the-art methods on MAVEN. The best results are highlighted in \textbf{bold} and the
second best results are \underline{underlined}.}
\label{tab:MAVEN_exp}
\centering
\resizebox{0.85\linewidth}{!}{
    \begin{tabular}{c|cc}
    \toprule
        \multirow{2}{*}{Methods}  & \multicolumn{2}{c}{MAVEN}  \\
        \cmidrule(r){2-3}
        & $\mathcal{A}_T$ & $\bar{\mathcal{A}}$ \\
         \midrule
        SEQ   & 3.69\tiny{ ± 0.17} & 11.75\tiny{ ± 0.13} \\
        SelfTrain & 35.33\tiny{ ± 0.41} & \underline{45.42\tiny{ ± 0.76}} \\
        ExtendNER & 13.81\tiny{ ± 0.56} & 24.92\tiny{ ± 0.74} \\
        CFNER & 22.74\tiny{ ± 1.52} & 34.77\tiny{ ± 1.38} \\
        DLD   & 14.18\tiny{ ± 0.37} & 24.98\tiny{ ± 0.43} \\
        RDP   & 28.76\tiny{ ± 2.44} & 38.01\tiny{ ± 1.09} \\
        OCILNER &  21.70\tiny{ ± 1.77}	& 30.13\tiny{ ± 0.75}\\
        ICE\_PLO & \underline{39.01\tiny{ ± 0.51}} & 44.02\tiny{ ± 0.96} \\
        ICE\_O & 38.16\tiny{ ± 1.26} & 43.43\tiny{ ± 1.31} \\
        CPFD  & 27.28\tiny{ ± 1.39} & 41.31\tiny{ ± 1.31} \\
        \midrule
        \rowcolor{black!10}\textbf{IS3 (Ours)} & \textbf{40.15\tiny{ ± 0.38}}&	\textbf{48.16\tiny{ ± 0.16}} \\ 
          \bottomrule
    \end{tabular}}
\end{table}

We compare the runtime and the number of updated parameters in Table \ref{tab:comparison_time_params}. The results indicate that the proposed IS3 requires less than 5\% more training time than SEQ and less training time than most previous methods, such as CFNER and CPFD.

\begin{table*}[htbp]
\caption{Comparisons with state-of-the-art methods on the OntoNotes5 dataset using the \textit{bert-base-cased} model. The best results are highlighted in \textbf{bold} and the
second best results are \underline{underlined}.}
\label{tab:basic_exp_OntoNotes5}
\centering
\resizebox{\textwidth}{!}{
    \begin{tabular}{cc|cc|cc|cc|cc}
    \toprule
        \multirow{2}{*}{Dataset} & \multirow{2}{*}{Methods}  & \multicolumn{2}{c}{FG-1-PG-1} \vline&  \multicolumn{2}{c}{FG-2-PG-2}   \vline & \multicolumn{2}{c}{FG-8-PG-1} \vline& \multicolumn{2}{c}{FG-8-PG-2}  \\
        \cmidrule(r){3-10}
        && $\mathcal{A}_T$ & $\bar{\mathcal{A}}$ & $\mathcal{A}_T$ & $\bar{\mathcal{A}}$ & $\mathcal{A}_T$ & $\bar{\mathcal{A}}$ & $\mathcal{A}_T$ & $\bar{\mathcal{A}}$ \\ 
         \midrule
        
        \multirow{9}{*}{OntoNotes5} & FT   & 1.65\tiny{ ± 0.11} & 12.91\tiny{ ± 0.41} & 4.49\tiny{ ± 0.44} & 20.69\tiny{ ± 0.25} & 1.42\tiny{ ± 0.08} & 12.41\tiny{ ± 0.38} & 3.97\tiny{ ± 0.37} & 21.45\tiny{ ± 0.28} \\
          & SelfTrain & 38.32\tiny{ ± 5.29} & 47.07\tiny{ ± 1.67} & \underline{52.23\tiny{ ± 0.43}} & \underline{56.14\tiny{ ± 0.88}} & 38.26\tiny{ ± 3.44} & 49.31\tiny{ ± 2.92} & \underline{51.71\tiny{ ± 1.39}} & 58.51\tiny{ ± 1.04} \\
          & ExtendNER & 28.62\tiny{ ± 2.42} & 42.20\tiny{ ± 2.16} & 45.05\tiny{ ± 0.61} & 52.30\tiny{ ± 1.03} & 25.71\tiny{ ± 5.67} & 40.34\tiny{ ± 3.64} & 44.82\tiny{ ± 2.42} & 55.25\tiny{ ± 1.58} \\
          & CFNER & \underline{44.76\tiny{ ± 0.28}} & \underline{50.76\tiny{ ± 1.61}} & 49.29\tiny{ ± 2.25} & 55.94\tiny{ ± 1.37} & \underline{46.81\tiny{ ± 0.99}} & \underline{54.91\tiny{ ± 0.69}} & 51.41\tiny{ ± 2.21} & \underline{60.41\tiny{ ± 0.43}} \\
          & DLD   & 22.22\tiny{ ± 5.38} & 38.47\tiny{ ± 4.73} & 44.88\tiny{ ± 0.78} & 51.91\tiny{ ± 1.15} & 25.25\tiny{ ± 1.69} & 41.43\tiny{ ± 1.01} & 44.53\tiny{ ± 1.66} & 55.17\tiny{ ± 1.18} \\
          & RDP   & 38.25\tiny{ ± 5.02} & 48.14\tiny{ ± 2.60} & 48.55\tiny{ ± 3.54} & 54.81\tiny{ ± 2.57} & 39.31\tiny{ ± 4.29} & 52.28\tiny{ ± 3.11} & 50.34\tiny{ ± 1.86} & 59.89\tiny{ ± 0.83} \\
          & OCILNER & 14.91\tiny{ ± 4.39} & 24.72\tiny{ ± 3.21} & 26.31\tiny{ ± 2.38} & 35.96\tiny{ ± 1.76} & 19.39\tiny{ ± 2.98} & 30.41\tiny{ ± 2.98} & 23.28\tiny{ ± 4.21} & 30.27\tiny{ ± 4.46} \\
          & ICE\_PLO & 39.69\tiny{ ± 0.36} & 43.76\tiny{ ± 0.16} & 43.81\tiny{ ± 0.34} & 46.38\tiny{ ± 0.36} & 42.69\tiny{ ± 0.09} & 46.95\tiny{ ± 0.21} & 44.66\tiny{ ± 0.61} & 47.72\tiny{ ± 0.61} \\
          & ICE\_O & 38.87\tiny{ ± 0.37} & 43.51\tiny{ ± 0.23} & 40.82\tiny{ ± 0.35} & 44.71\tiny{ ± 0.28} & 45.98\tiny{ ± 0.28} & 49.11\tiny{ ± 0.49} & 48.01\tiny{ ± 0.49} & 49.91\tiny{ ± 0.57} \\
          & CPFD  & 33.44\tiny{ ± 1.18} & 44.73\tiny{ ± 0.69} & 43.48\tiny{ ± 0.72} & 50.79\tiny{ ± 1.05} & 41.77\tiny{ ± 2.79} & 52.46\tiny{ ± 1.02} & 48.36\tiny{ ± 2.35} & 58.60\tiny{ ± 1.99} \\

        \midrule
        \rowcolor{black!10}&\textbf{IS3 (Ours)} & \textbf{50.23\tiny{ ± 0.94}}&	\textbf{54.65\tiny{ ± 0.84}}&	\textbf{57.23\tiny{ ± 1.19}}&	\textbf{58.25\tiny{ ± 0.56}}&	\textbf{56.11\tiny{ ± 1.15}}&	\textbf{61.44\tiny{ ± 0.11}}&	\textbf{62.23\tiny{ ± 0.10}}&	\textbf{66.01\tiny{ ± 0.74}}\\ 
          \bottomrule
    \end{tabular}}
\end{table*}

\begin{table*}[htbp]
\caption{Comparisons with state-of-the-art methods on the i2b2 dataset using the \textit{roberta-base} model. The best results are highlighted in \textbf{bold} and the
second best results are \underline{underlined}. The average of each incremental step is provided in Figure \ref{fig:line_graph}.}
\label{tab:basic_exp_i2b2_robert}
    \centering
\resizebox{\textwidth}{!}{
    \begin{tabular}{cc|cc|cc|cc|cc}
    \toprule
          \multirow{2}{*}{Dataset} &\multirow{2}{*}{Methods}  & \multicolumn{2}{c}{FG-1-PG-1} \vline&  \multicolumn{2}{c}{FG-2-PG-2}   \vline & \multicolumn{2}{c}{FG-8-PG-1} \vline& \multicolumn{2}{c}{FG-8-PG-2}  \\
        \cmidrule(r){3-10}
        & $\mathcal{A}_T$ & $\bar{\mathcal{A}}$ & $\mathcal{A}_T$ & $\bar{\mathcal{A}}$ & $\mathcal{A}_T$ & $\bar{\mathcal{A}}$ & $\mathcal{A}_T$ & $\bar{\mathcal{A}}$ \\ 
         \midrule
        \multirow{9}{*}{i2b2} & FT & 2.68\tiny{ ± 1.19} & 14.36\tiny{ ± 0.81} & 7.39\tiny{ ± 1.58} & 23.02\tiny{ ± 0.42} & 1.91\tiny{ ± 0.38} & 15.56\tiny{ ± 1.77} & 6.00\tiny{ ± 0.49} & 25.01\tiny{ ± 0.73} \\
        & SelfTrain & 17.58\tiny{ ± 1.60} & 37.95\tiny{ ± 1.10} & 25.84\tiny{ ± 3.26} & 43.80\tiny{ ± 3.24} & 6.97\tiny{ ± 1.15} & 28.73\tiny{ ± 0.48} & 27.85\tiny{ ± 3.72} & 45.80\tiny{ ± 2.37} \\
        & ExtendNER & 17.68\tiny{ ± 1.75} & 34.53\tiny{ ± 2.58} & 26.84\tiny{ ± 1.63} & 44.33\tiny{ ± 3.21} & 9.49\tiny{ ± 1.07} & 28.80\tiny{ ± 0.72} & 22.27\tiny{ ± 7.08} & 40.07\tiny{ ± 5.62} \\
        & CFNER & \underline{32.65\tiny{ ± 1.87}} & \underline{47.06\tiny{ ± 2.70}} & \underline{43.12\tiny{ ± 2.84}} & \underline{54.61\tiny{ ± 2.40}} & 33.52\tiny{ ± 0.78} & 38.61\tiny{ ± 1.56} & 36.19\tiny{ ± 7.72} & 49.46\tiny{ ± 6.09} \\
        & DLD   & 16.26\tiny{ ± 4.79} & 34.02\tiny{ ± 3.46} & 27.12\tiny{ ± 7.30} & 45.81\tiny{ ± 3.26} & 5.54\tiny{ ± 1.68} & 27.80\tiny{ ± 1.39} & 20.39\tiny{ ± 4.46} & 38.43\tiny{ ± 2.58} \\
        & RDP   & 21.70\tiny{ ± 0.88} & 40.71\tiny{ ± 2.86} & 35.38\tiny{ ± 2.79} & 53.81\tiny{ ± 0.98} & 26.49\tiny{ ± 6.16} & 39.54\tiny{ ± 3.96} & \underline{42.98 \tiny{ ± 4.06}} & \underline{56.25\tiny{ ± 1.44}} \\
        & OCILNER & 9.27\tiny{ ± 4.60} & 27.98\tiny{ ± 1.94} & 15.14\tiny{ ± 9.85} & 34.44\tiny{ ± 4.11} & 15.89\tiny{ ± 8.39} & 35.54\tiny{ ± 7.28} & 21.26\tiny{ ± 8.70} & 39.21\tiny{ ± 5.01} \\
        & ICE\_PLO & 29.89\tiny{ ± 0.23} & 36.65\tiny{ ± 0.32} & 32.14\tiny{ ± 0.36} & 37.34\tiny{ ± 0.31} & \underline{34.34\tiny{ ± 0.84}} & \underline{41.07\tiny{ ± 0.91}} & 39.88\tiny{ ± 0.71} & 44.34\tiny{ ± 0.46} \\
        & ICE\_O & 25.56\tiny{ ± 0.94} & 35.57\tiny{ ± 0.75} & 33.17\tiny{ ± 0.27} & 38.16\tiny{ ± 0.25} & 32.77\tiny{ ± 1.16} & 40.58\tiny{ ± 1.30} & 36.95\tiny{ ± 1.41} & 41.95\tiny{ ± 1.44} \\
        & CPFD  & 13.65\tiny{ ± 4.16} & 40.81\tiny{ ± 3.10} & 26.38\tiny{ ± 1.72} & 49.82\tiny{ ± 1.41} & 4.59\tiny{ ± 0.86} & 31.76\tiny{ ± 2.39} & 25.23\tiny{ ± 9.73} & 48.85\tiny{ ± 3.37} \\
        \midrule
        \rowcolor{black!10}&\textbf{IS3 (Ours)}  & \textbf{34.14\tiny{ ± 1.45}} & \textbf{51.61\tiny{ ± 1.60}} & \textbf{47.91\tiny{ ± 2.76}} & \textbf{59.22\tiny{ ± 1.01}} & \textbf{42.99\tiny{ ± 3.01}} & \textbf{52.25\tiny{ ± 1.96}} & \textbf{52.36\tiny{ ± 3.42}} & \textbf{60.99\tiny{ ± 1.49}} \\	 
        \bottomrule
    \end{tabular}}
\end{table*}

\begin{table*}[htbp]
\caption{Comparisons with state-of-the-art methods on the OntoNotes5 dataset using the \textit{roberta-base} model. The best results are highlighted in \textbf{bold} and the
second best results are \underline{underlined}.}
\label{tab:basic_exp_OntoNotes5_robert}
    \centering
\resizebox{\textwidth}{!}{
    \begin{tabular}{cc|cc|cc|cc|cc}
    \toprule
        \multirow{2}{*}{Dataset} & \multirow{2}{*}{Methods}  & \multicolumn{2}{c}{FG-1-PG-1} \vline&  \multicolumn{2}{c}{FG-2-PG-2}   \vline & \multicolumn{2}{c}{FG-8-PG-1} \vline& \multicolumn{2}{c}{FG-8-PG-2}  \\
        \cmidrule(r){3-10}
        && $\mathcal{A}_T$ & $\bar{\mathcal{A}}$ & $\mathcal{A}_T$ & $\bar{\mathcal{A}}$ & $\mathcal{A}_T$ & $\bar{\mathcal{A}}$ & $\mathcal{A}_T$ & $\bar{\mathcal{A}}$ \\ 
         \midrule
        
        \multirow{9}{*}{OntoNotes5} &FT    & 1.75\tiny{ ± 0.11} & 12.43\tiny{ ± 0.11} & 5.07\tiny{ ± 0.32} & 20.85\tiny{ ± 0.27} & 1.32\tiny{ ± 0.26} & 12.03\tiny{ ± 0.20} & 4.97\tiny{ ± 0.33} & 20.82\tiny{ ± 0.14} \\
        & SelfTrain & 38.88\tiny{ ± 6.38} & 47.57\tiny{ ± 2.30} & 54.02\tiny{ ± 0.76} & 55.78\tiny{ ± 1.87} & 39.82\tiny{ ± 5.23} & 49.05\tiny{ ± 2.34} & 51.77\tiny{ ± 2.28} & 57.71\tiny{ ± 0.48} \\
        & ExtendNER & 23.29\tiny{ ± 5.15} & 36.77\tiny{ ± 6.26} & 44.00\tiny{ ± 2.82} & 50.06\tiny{ ± 1.44} & 24.12\tiny{ ± 1.85} & 40.39\tiny{ ± 3.00} & 44.48\tiny{ ± 5.62} & 53.50\tiny{ ± 2.86} \\
        & CFNER & \underline{41.86\tiny{ ± 2.78}} & \underline{48.73\tiny{ ± 3.11}} & \underline{54.24\tiny{ ± 1.04}} & \underline{58.07\tiny{ ± 1.71}} & \underline{51.51\tiny{ ± 2.06}} & \underline{55.65\tiny{ ± 0.56}} & 52.23 \tiny{± 2.75} & 58.95\tiny{ ± 4.11} \\
        & DLD   & 25.25\tiny{ ± 4.17} & 37.24\tiny{ ± 6.06} & 46.58\tiny{ ± 1.78} & 50.93\tiny{ ± 0.54} & 24.04\tiny{ ± 2.54} & 40.63\tiny{ ± 0.91} & 45.89\tiny{ ± 1.27} & 54.89\tiny{ ± 0.38} \\
        & RDP   & 34.84\tiny{ ± 2.92} & 45.90\tiny{ ± 0.91} & 46.95\tiny{ ± 1.49} & 55.02\tiny{ ± 0.73} & 41.58\tiny{ ± 3.37} & 53.69\tiny{ ± 0.53} & \underline{52.54\tiny{ ± 0.44}} & \underline{62.03\tiny{ ± 1.05}} \\
        & OCILNER & 27.98\tiny{ ± 5.18} & 29.86\tiny{ ± 4.28} & 29.64\tiny{ ± 4.27} & 37.24\tiny{ ± 2.79} & 29.94\tiny{ ± 1.60} & 42.51\tiny{ ± 1.13} & 30.90\tiny{ ± 2.41} & 45.54\tiny{ ± 0.63} \\
        & ICE\_PLO & 33.66\tiny{ ± 0.66} & 37.21\tiny{ ± 0.36} & 36.17\tiny{ ± 0.07} & 37.71\tiny{ ± 0.59} & 36.90\tiny{ ± 0.19} & 40.33\tiny{ ± 0.37} & 38.83\tiny{ ± 0.34} & 41.10\tiny{ ± 0.30} \\
        & ICE\_O & 30.46\tiny{ ± 1.69} & 33.41\tiny{ ± 1.91} & 35.05\tiny{ ± 0.92} & 37.17\tiny{ ± 1.12} & 37.38\tiny{ ± 1.14} & 40.22\tiny{ ± 0.83} & 40.69\tiny{ ± 0.21} & 42.25\tiny{ ± 0.47} \\
        & CPFD  & 29.50\tiny{ ± 1.47} & 43.12\tiny{ ± 2.18} & 41.01\tiny{ ± 2.92} & 51.56\tiny{ ± 0.52} & 40.58\tiny{ ± 2.50} & 51.64\tiny{ ± 2.52} & 50.70\tiny{ ± 3.65} & 61.14\tiny{ ± 1.57} \\
        \midrule
        \rowcolor{black!10}&\textbf{IS3 (Ours)} & \textbf{42.69\tiny{ ± 1.21}}&	\textbf{49.73\tiny{ ± 1.13}}&	\textbf{58.77\tiny{ ± 1.97}}&	\textbf{59.27\tiny{ ± 1.28}}&	\textbf{55.43\tiny{ ± 2.01}}&	\textbf{60.25\tiny{ ± 1.49}}&	\textbf{61.75\tiny{ ± 0.96}}&	\textbf{64.61\tiny{ ± 0.78}}\\ 
          \bottomrule
    \end{tabular}}
\end{table*}

\end{document}